\documentclass[review,a4paper]{elsarticle}

\usepackage{geometry}
\usepackage{bm}
\usepackage{float}
\usepackage{stfloats}
\usepackage{booktabs}
\usepackage{lscape}

\usepackage{siunitx}
\usepackage{adjustbox}
\usepackage{amssymb}  
\usepackage{enumitem} 
\usepackage{amsmath} 
\usepackage{hyperref}
\usepackage{multirow}
\usepackage[figuresright]{rotating}
\usepackage{longtable}
\usepackage{caption}
\usepackage{graphicx}
\usepackage{pdflscape} 

\journal{Journal of \LaTeX\ Templates}

\usepackage[linesnumbered,ruled,vlined]{algorithm2e}
\usepackage{subcaption} 
\usepackage{pgffor} 
\usepackage[table]{xcolor} 

\begin{document}

\begin{frontmatter}

\title{LLMDE: A Large Language Model-Driven Differential Evolution Algorithm for Portfolio Optimization} 

\author[1]{Rong Chai}

\author[1]{V\'{a}clav Sn\'{a}\v{s}el}

\author[1]{Xiaopeng Wang}

\author[1,2]{Seyedali Mirjalili\corref{correspondingauthor}}
\ead{ali.mirjalili@gmail.com}

\author[3]{Crina Grosan}

\cortext[correspondingauthor]{Corresponding author}

\address[1]{Faculty of Electrical Engineering and Computer Science, V\v{S}B-Technical University of Ostrava, Ostrava, Czech Republic} 
\address[2]{Centre for Artificial Intelligence Research and Optimisation, Torrens University Australia, Brisbane, Australia}
\address[3]{Digital Health and Applied Technology Assessment, King’s College London, London, UK}

\begin{abstract}
This study proposes a Large Language Model-Driven Differential Evolution (LLMDE) algorithm to reduce the reliance on handcrafted hyperparameter design. The proposed algorithm leverages a prompt engineering strategy, allowing large language models (LLMs) to dynamically select mutation strategies and configure control parameters guided by optimization feedback, thus enhancing the performance of the DE algorithm. We evaluate the performance of LLMDE on the CEC2022 benchmark suite, comparing it with standard DE and representative metaheuristics. Furthermore, we employ factor analysis and K-means clustering for stock selection, and then apply LLMDE to solve the Conditional Value at Risk (CVaR) portfolio optimization problem using the selected stocks, subject to budget and minimum expected return constraints. Experimental results demonstrate that LLMDE achieves competitive performance on the benchmark suite while continuously generating high-quality solutions for complex constrained optimization tasks. These outcomes successfully demonstrate the viability of embedding LLMs within metaheuristics, paving a promising path toward the design of advanced LLM-assisted optimization techniques. 

\end{abstract}


\end{frontmatter}


\section{Introduction}

Portfolio optimization is a fundamental problem in quantitative finance, aiming to achieve the optimal trade-off between risk and return by maximizing the expected return for a given level of risk or minimizing the portfolio risk for a specified target return. By appropriately allocating investment weights across assets, investors can exploit correlations among asset returns to diversify risk and thereby improve risk-adjusted returns \cite{gunjan2023brief}. Modern Portfolio Theory (MPT), pioneered by Harry Markowitz in 1952 \cite{Markowitz}, laid the theoretical foundation for contemporary portfolio management. Crucial to this theory is the mean–variance (MV) framework, which formalizes return and risk via expected return and variance, a paradigm that inherently assumes normally distributed asset returns. However, real-world financial returns typically violate this assumption by exhibiting distinct skewness, kurtosis, and heavy tails. In such cases, variance is insufficient to accurately capture downside risk and potential extreme losses, thereby limiting the effectiveness of the traditional MV framework in practical financial risk management.

To effectively capture extreme losses, Value at Risk (VaR) has been widely employed to quantify the maximum potential portfolio loss over a given time horizon at a specific confidence level \cite{jorion2007value}. However, because VaR merely reflects a distinct quantile of the loss distribution, it provides no insight into the magnitude of losses beyond this benchmark, thus failing to adequately account for catastrophic tail events. In light of these limitations, Rockafellar and Uryasev pioneered Conditional Value-at-Risk (CVaR), which computes the expected loss in the tail of the distribution exceeding the VaR threshold \cite{rockafellar2000optimization}. As a coherent risk measure, CVaR has become a cornerstone of modern portfolio optimization, with widespread implementation across asset allocation, risk management, and financial regulation \cite{krokhmal2002portfolio}. Despite CVaR's superior risk-measuring capabilities, CVaR-based portfolio optimization is highly challenging due to non-linearity and computational complexity. Incorporating realistic constraints like cardinality, budget, and minimum target returns typically classifies the formulation as NP-hard, especially as the asset count scales. 

Metaheuristic algorithms have become a promising alternative for solving complex optimization problems owing to their flexibility, robustness, and strong global search capabilities. Among these approaches, Differential Evolution (DE) \cite{price2006differential} has received extensive attention due to its simplicity, rapid convergence, and demonstrated effectiveness in continuous optimization. Consequently, DE has been successfully employed across a variety of engineering and financial optimization tasks. Nevertheless, the performance of DE is highly sensitive to the choice of mutation strategies and the configuration of control parameters, specifically the mutation factor ($F$) and crossover rate $(\mathit{CR})$. Improper parameter settings can trigger premature convergence or a severe loss of population diversity, ultimately compromising the algorithm’s search efficiency and solution quality.

Large Language Models (LLMs) are driving a paradigm shift from expert-driven algorithm design to automated optimization frameworks through their capacity for automatic generation and adaptive refinement. 
Leveraging their advanced capabilities in information processing and contextual reasoning, LLMs offer a compelling avenue for dynamically orchestrating search strategies in response to the evolving optimization state. Existing studies have explored the use of LLMs in optimization from multiple perspectives. Zhong et al. (2024) \cite{zhong2024leveraging} employed ChatGPT with the CRISPE prompt engineering framework to design a novel animal-inspired metaheuristic algorithm, Zoological Search Optimization (ZSO), based on prey–predator interaction and flocking behaviors. Qi et al. (2025) \cite{qi2025memetic} proposed a memetic framework integrating LLM-driven exploration, local optimization, and memory-aware reflection for AGV-drone scheduling and combinatorial optimization problems, demonstrating the effectiveness of LLMs as interpretable heuristic generators. Sartori and Blum's research (2025) \cite{sartori2025combinatorial} focused on using LLMs to directly improve existing algorithm codebases without requiring expert knowledge, achieving significant improvements in 9 out of 10 baseline algorithms for the Traveling Salesman Problem. Wang et al. (2026) \cite{wang2026towards}  developed an LLM-driven automatic genetic algorithm design framework for weather routing, enabling the adaptive generation of crossover and mutation operators tailored to specific voyage conditions. Zhang et al. (2026) \cite{zhang2026budget} proposed an LLM-guided ant colony optimization method for wireless energy service composition, using reflective evolutionary feedback to model user preferences and guide the search process. Carr et al. (2026) \cite{carr2026automated} proposed an LLM-guided evolutionary framework for robot control systems, separating control logic synthesis and numerical optimization for UAV trajectory tracking, achieving at least a 38\%  reduction in mean squared error in experiments.

Although LLMs have shown immense potential in algorithm development, existing research primarily targets automated code generation and offline heuristic synthesis. In contrast, leveraging LLMs to adaptively guide evolutionary search using real-time optimization feedback remains an open challenge, and its application to complex CVaR portfolio optimization is still remarkably scarce. The No Free Lunch theorem \cite{wolpert1997no} dictates that no single optimization algorithm can universally outperform all others. Consequently, developing adaptive mechanisms that dynamically adjust search strategies based on real-time feedback is crucial for enhancing algorithmic efficacy. To address this, we introduce a Large Language Model-driven Differential Evolution (LLMDE) algorithm, which integrates an LLM into the DE framework to adaptively control the evolutionary search process. The primary contributions of this study are summarized as follows.

\begin{enumerate}
\item The proposed LLMDE algorithm integrates an LLM into DE for adaptive search control, thereby eliminating the reliance on manually designed mechanisms.
\item The LLM dynamically configures mutation strategies and control parameters during the optimization process, achieving a more effective balance between exploration and exploitation.
\item Extensive experiments on the CEC 2022 benchmark suite and a real-world CVaR portfolio optimization task validate LLMDE’s superior capability in solving complex optimization problems. 
\end{enumerate}

The rest of this paper is structured as follows. Section 2 outlines the foundational principles and computational complexity of DE, followed by the mathematical formulation of the CVaR portfolio optimization problem. Section 3 introduces the proposed LLMDE framework and its key mechanisms. Section 4 validates its algorithmic performance on the CEC2022 benchmark functions, while Section 5 showcases its application to real-world CVaR portfolio optimization. Finally, Section 6 summarizes this study and highlights avenues for future work.

\section{Preliminaries}

This section presents the theoretical background of the proposed method: the classical DE algorithm as the optimizer, and the CVaR portfolio model as the optimization objective.

\subsection{Differential Evolution}

Differential Evolution (DE) \cite{price2006differential} is a powerful population-based, gradient-free stochastic optimization method. The algorithm navigates complex search spaces by iteratively refining a population of candidate solutions through distinct phases of mutation, crossover, and selection. This exceptional efficacy makes DE highly effective in optimizing non-linear and rugged landscapes. The core operational steps of the DE algorithm are detailed below.

\textbf{Initialization:} To ensure broad population diversity, an initial population of $N$ individuals is randomly generated within the predefined bounds of the $D$-dimensional search space. Specifically, each individual's position vector is sampled from a uniform distribution across the entire search domain.

\begin{equation}
\mathbf{x}_i = \mathbf{x}_\mathrm{min} + \mathbf{r}_i \odot (\mathbf{x}_\mathrm{max} - \mathbf{x}_\mathrm{min})
\label{Eq1}
\end{equation}
where:
\begin{itemize} [label=\textbullet, itemsep=6pt, topsep=3pt, leftmargin=18pt]
    \item The individual vector is: $\mathbf{x}_i = [x_{i,1}, x_{i,2}, \dots, x_{i,D}]^T$.
    \item The lower boundary vector is: $\mathbf{x}_\mathrm{min} = [x_1^\mathrm{min}, x_2^\mathrm{min}, \dots, x_D^\mathrm{min}]^T$. 
    \item The upper boundary vector is: $\mathbf{x}_\mathrm{max} = [x_1^\mathrm{max}, x_2^\mathrm{max}, \dots, x_D^\mathrm{max}]^T$.
    \item The random vector is defined as: $\mathbf{r}_i = [r_{i,1}, r_{i,2}, \dots, r_{i,D}]^T, \quad r_{i,j} \sim U(0,1)$.
    \item $\odot$ denotes the element-wise (Hadamard) product.
\end{itemize}

\textbf{Mutation:} Following initialization, the mutation operator is applied to generate a donor (mutant) vector $\mathbf{v}_i$ for each target individual $\mathbf{x}_i$. This is achieved by adding one or more weighted difference vectors to a designated base vector. In the standard nomenclature denoted as DE/$x$/$y$/$z$, $x$ specifies the base vector selection strategy, $y$ indicates the number of difference vectors employed, and $z$ represents the subsequent crossover scheme. To balance exploration and exploitation, various mutation configurations can be deployed; Table~\ref{Tab1} summarizes five widely adopted mutation operators. Among them, the foundational mathematical formulation for DE/rand/1 is defined as follows:

\begin{equation}
\mathbf{v}_i = \mathbf{x}_{r_1} + F \cdot (\mathbf{x}_{r_2} - \mathbf{x}_{r_3})
\label{Eq2}
\end{equation}
where:
\begin{itemize} [label=\textbullet, itemsep=6pt, topsep=3pt, leftmargin=18pt]
    \item $\mathbf{x}_{r_1}, \mathbf{x}_{r_2}$, and $\mathbf{x}_{r_3}$ are three distinct individuals randomly selected from the current population.
    \item $F$ is the scaling factor.
\end{itemize}

\renewcommand\arraystretch{1.5}
\begin{table}[h] \scriptsize
    \caption{DE mutation operators.}
    \begin{center}
    \begin{tabular}{>{\centering\arraybackslash}p{1.5cm} >{\raggedright\arraybackslash}p{3.6cm} >{\raggedright\arraybackslash}p{6.5cm}}
    \toprule
    \textbf{No.} & \textbf{Operator} & \textbf{Equation} \\ 
    \midrule
    0. & DE/rand/2 & $\mathbf{v}_i = \mathbf{x}_{r1} + F \cdot (\mathbf{x}_{r2} - \mathbf{x}_{r3}) + F \cdot (\mathbf{x}_{r4} - \mathbf{x}_{r5})$ \\ 
    1. & DE/rand/1 & $\mathbf{v}_i = \mathbf{x}_{r1} + F \cdot (\mathbf{x}_{r2} - \mathbf{x}_{r3})$ \\ 
    2. & DE/current-to-best/1 & $\mathbf{v}_i = \mathbf{x}_i + F \cdot (\mathbf{x}_\mathrm{best} - \mathbf{x}_i) + F \cdot (\mathbf{x}_{r1} - \mathbf{x}_{r2})$ \\ 
    3. & DE/best/2 & $\mathbf{v}_i = \mathbf{x}_\mathrm{best} + F \cdot (\mathbf{x}_{r1} - \mathbf{x}_{r2}) + F \cdot (\mathbf{x}_{r3} - \mathbf{x}_{r4})$ \\ 
    4. & DE/best/1 & $\mathbf{v}_i = \mathbf{x}_\mathrm{best} + F \cdot (\mathbf{x}_{r1} - \mathbf{x}_{r2})$ \\ 
    \bottomrule
    \end{tabular}
    \end{center}
    \label{Tab1}
\end{table}

\textbf{Crossover:} Upon generating the donor vector, a crossover operation is executed to construct the trial vector $\mathbf{u}_i$ by recombining the target vector $\mathbf{x}_i$ and the donor vector $\mathbf{v}_i$. This study utilizes the widely adopted binomial crossover scheme, which cross-references genetic components based on a predefined crossover probability $\mathit{CR} \in [0, 1]$. To prevent stagnation and ensure that the trial vector inherits at least one component from the mutant vector, a randomly selected dimension is forcibly copied. Formally, the trial vector $\mathbf{u}_i$ is defined as follows:

\begin{equation}
\mathbf{u}_i = \mathbf{x}_i \odot (\mathbf{1} - \mathbf{c}_i) + \mathbf{v}_i \odot \mathbf{c}_i
\label{Eq3}
\end{equation}
where:
\begin{itemize} [label=\textbullet, itemsep=6pt, topsep=3pt, leftmargin=18pt]
    \item $\mathbf{u}_i, \mathbf{x}_i, \mathbf{v}_i \in \mathbb{R}^D$ are the trial, target, and donor vectors, respectively.
    \item $\mathbf{1}$ denotes a vector of ones with length $D$.
    \item $\mathbf{c}_i = [c_{i,1}, c_{i,2}, \dots, c_{i,D}]^T$,
            \[ 
            c_{i,j} =
            \begin{cases}
            1, & \text{if } rand_j \leq \textit{CR} \text{ or } j = j_\mathrm{rand} \\
            0, & \text{otherwise}
            \end{cases}
            \]
    \item \textit{CR} is the crossover rate in $[0,1]$, and $j_\mathrm{rand}$ is a randomly selected index in $[1,D]$.
\end{itemize}

\textbf{Selection:} To determine survival into the subsequent generation, a greedy selection mechanism is executed by comparing each trial vector $\mathbf{u}_i$ against its corresponding target vector $\mathbf{x}_i$. The vector yielding the superior fitness value is preserved for the next iteration, while the inferior one is eliminated. This paradigm maintains a constant population size while driving convergence toward global optima. Mathematically, the selection operator is expressed as:

\begin{equation}
\mathbf{x}_i =
    \begin{cases}
    \mathbf{u}_i, & \text{if } f(\mathbf{u}_i) \leq f(\mathbf{x}_i) \\
    \mathbf{x}_i, & \text{otherwise}
    \end{cases}
\label{Eq4}
\end{equation}

\subsection{Characteristics and Complexity Analysis of DE}

As an effective population-based metaheuristic, Differential Evolution (DE) is characterized by its structural simplicity and gradient-free nature. By utilizing difference vectors to direct individual variations, DE exhibits notable robustness in addressing non-linear, non-differentiable, and complex black-box landscapes with minimal memory overhead. However, DE’s performance relies heavily on proper operator and parameter configurations. Moreover, severe diversity loss in later generations frequently induces search stagnation and slow convergence, increasing its risk of trapping in local optima on complex multimodal landscapes.

The time complexity of DE is governed by the population size ($N$), problem dimension ($D$), maximum number of generations ($T$), and the evaluation complexity of the objective function ($f_{\mathrm{eval}}$). Within each generation, the algorithm sequentially executes mutation, crossover, and selection. Because mutation and crossover are performed dimension by dimension while selection requires pairwise fitness evaluations and solution updates, all three operators operate linearly with respect to population size and dimension, yielding an intrinsic per-generation time complexity of $\mathcal{O}(N \cdot D)$. Incorporating the complete evolutionary cycle and objective function evaluations, the total time complexity is formulated as $\mathcal{O}(T \cdot N \cdot (D + f_{\mathrm{eval}}))$.

Regarding memory complexity, DE primarily maintains the candidate population and its associated fitness values. The runtime memory overhead is dominated by an $N \times D$ population matrix alongside an $\mathcal{O}(N)$ array of objective values, yielding an overall memory complexity of $\mathcal{O}(N \cdot D)$. Because DE routinely updates the population in place without archiving historical iterations, its spatial requirement remains independent of the generation budget $T$. Consequently, DE demonstrates notable memory efficiency during execution.

\subsection{CVaR Portfolio Optimization Model}

Conditional Value at Risk (CVaR) characterizes the average loss exceeding the VaR threshold \cite{rockafellar2000optimization}. As a coherent risk measure, the CVaR of a portfolio at the confidence level $\alpha$ is defined as:
\begin{equation}
CVaR_{\alpha}(\mathbf{x}) = E[f(\mathbf{x}, \mathbf{y}) \mid f(\mathbf{x}, \mathbf{y}) \geq VaR_{\alpha}(\mathbf{x})]=
\frac{1}{1-\alpha}
\int_{f(\mathbf{x},\mathbf{y})\ge VaR_{\alpha}(\mathbf{x})}
f(\mathbf{x},\mathbf{y})\,p(\mathbf{y})\,d\mathbf{y}
\label{Eq5}
\end{equation}

Since the formulation is difficult to optimize directly, the reformulation proposed by Rockafellar and Uryasev is adopted.
\begin{equation}
F_{\alpha}(\mathbf{x},VaR)
=
VaR
+
\frac{1}{1-\alpha}
\int
\left[f(\mathbf{x},\mathbf{y})-VaR\right]^+
p(\mathbf{y})\,d\mathbf{y}
\label{Eq6}
\end{equation}
where 
$\mathbf{x}=[x_1,x_2,\ldots,x_n]^T$ denotes the portfolio weight vector,
$\mathbf{y}$ is the return vector of assets,
$f(\mathbf{x},\mathbf{y})$ is the portfolio loss function,
$p(\mathbf{y})$ denotes the probability density function of $\mathbf{y}$,
$\alpha\in(0,1)$ is the confidence level,
and the positive-part operator is defined as
\[
\left[f(\mathbf{x},\mathbf{y})-VaR\right]^+
=
\max\left\{f(\mathbf{x},\mathbf{y})-VaR,\;0\right\}
\]

Accordingly, the CVaR portfolio optimization model can be formulated as follows:
\begin{equation}
\begin{aligned}
\min\quad
& F_{\alpha}(\mathbf{x}, VaR) 
= VaR + \frac{1}{(1-\alpha)S} 
\sum_{j=1}^{S} \max\{0, -\mathbf{x}^T \mathbf{y}_j - VaR\} \\
\text{s.t.} \quad 
& \begin{cases} 
\mathbf{x}^T \mathbf{r} \ge  r_{\mathrm{target}}, \\
\mathbf{1}^T \mathbf{x} = 1, \\
0 \le x_i \le 1, \quad i = 1, \dots, n
\end{cases}
\end{aligned}
\label{Eq7}
\end{equation}
where $\mathbf{y}_j$ denotes the asset return vector under the $j$th scenario, 
$\mathbf{r}$ denotes the expected return vector of the assets, 
$S$ is the number of scenarios (sample size), 
$r_{\mathrm{target}}$ is the target return, and 
$n$ is the number of assets.

For the multi-constraint CVaR model, an exterior penalty function method is adopted to convert the constraints into penalty terms, which are subsequently embedded into the objective function to construct an augmented objective formulation. This transformation unifies both feasible and infeasible regions within a single fitness landscape, thereby enabling metaheuristics to operate without explicit constraint-handling mechanisms. Consequently, the algorithm searches for high-quality solutions solely through objective function evaluation. The transformed objective function is formulated as follows:
\begin{equation}
\min\quad F_{\alpha}(\mathbf{x}, VaR)  = VaR + \frac{1}{(1-\alpha)S} \sum_{j=1}^{S} \max\{0, -\mathbf{x}^T \mathbf{y}_j - VaR\} + \sigma_1 \cdot \max(0, r_{\mathrm{target}}  - \mathbf{x}^T \mathbf{r}) + \sigma_2 \cdot |\mathbf{1}^T \mathbf{x} - 1|
\label{Eq8}
\end{equation}
where $\sigma_1$ and $\sigma_2$ are penalty factors.

\section{The Proposed LLMDE Algorithm}

This section details the proposed LLMDE algorithm. First, the overall framework is outlined to illustrate the interaction between Differential Evolution and the Large Language Model. Subsequently, the LLM prompt design is introduced, followed by a computational complexity analysis.

\subsection{The LLMDE Framework} 

The performance of Differential Evolution (DE) critically hinges on the configuration of mutation strategies and control parameters. Although conventional adaptive DE variants incorporate hand-designed adaptation mechanisms to maintain population diversity and prevent premature convergence, their efficacy remains vulnerable to strategy-state misalignment. Once an unsuitable strategy is deployed, the population quickly loses diversity and becomes trapped in local search spaces, especially when navigating complex multimodal functions. Furthermore, hand-crafting these adaptive rules requires extensive human expertise and trial-and-error tuning, which fundamentally restricts their adaptability and generalization across diverse problem domains. To circumvent these limitations and enable autonomous, context-aware search control, this study introduces a Large Language Model-driven Differential Evolution (LLMDE) algorithm.

Figure~\ref{Fig1} illustrates the systematic framework of LLMDE. The architecture comprises two core components: a Differential Evolution optimization engine and a Large Language Model (LLM)-driven strategy scheduling module, which interact through a closed-loop feedback mechanism. The DE module executes fundamental evolutionary operations, including population initialization, fitness evaluation, and generation updating. Concurrently, the LLM module analyzes critical population metrics and dynamically assigns suitable mutation strategies and control parameters to guide the search process. The detailed procedural steps are outlined below:

\begin{figure}[!t]
    \centering
    \includegraphics[width=\textwidth]{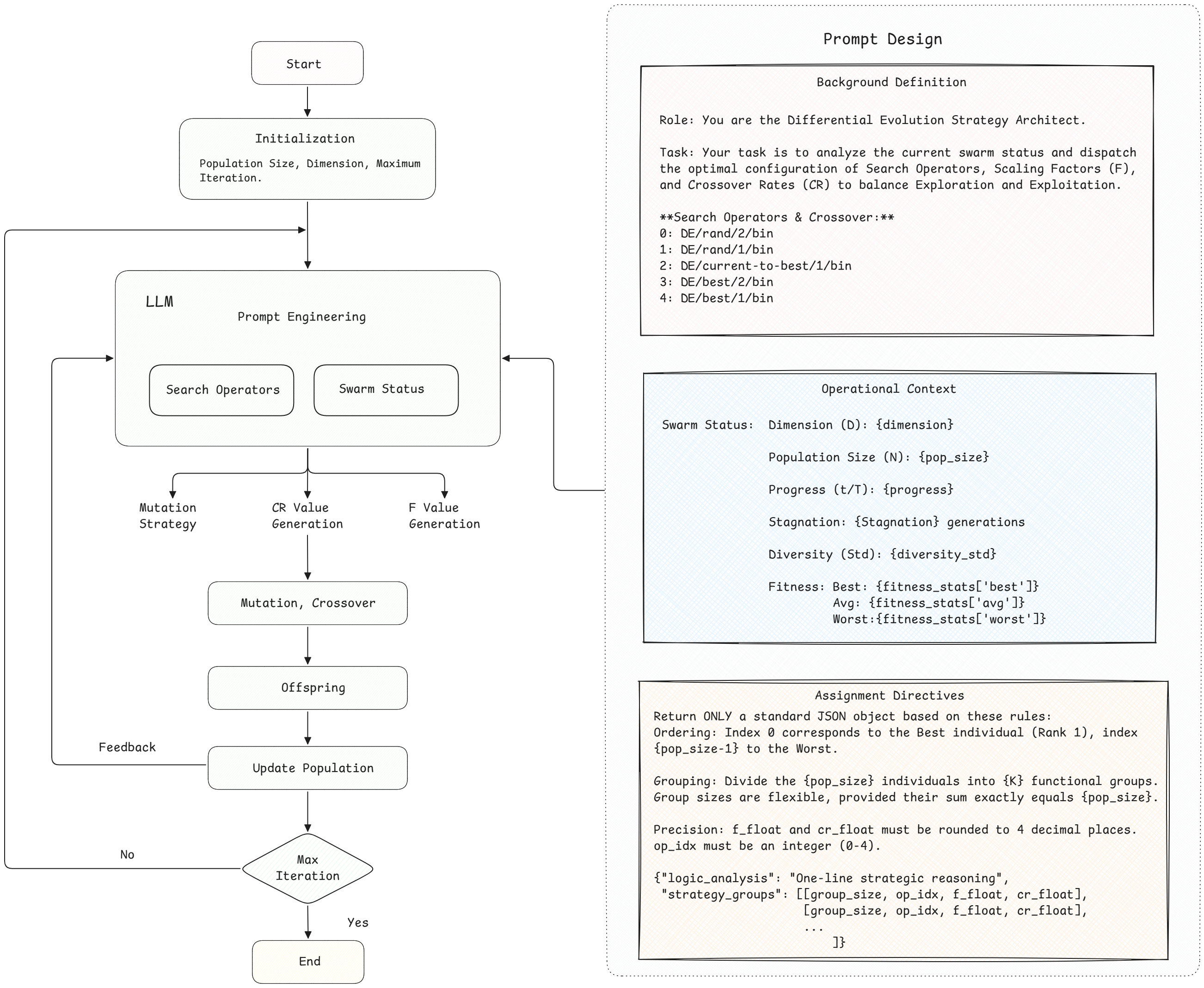}
    \caption{Systematic framework of the LLMDE algorithm, illustrating the interaction between the DE optimization workflow and the LLM prompt engineering design.}
    \label{Fig1}
\end{figure}

\begin{itemize}
    \item \textbf{Population Initialization and Status Extraction}:  
    LLMDE initializes the search process by generating a random population according to Eq.~\ref{Eq1} and evaluating individual fitness. Throughout the evolutionary process, the algorithm systematically extracts key swarm status descriptors, including search progress metrics, stagnation indicators, population diversity features, and fitness distribution statistics. These state descriptors, combined with the candidate search operator pool, are subsequently fed into the LLM as contextual inputs.
    
    \item \textbf{LLM-Guided Strategy Scheduling and Evolution}:  
    Guided by prompt engineering, the LLM analyzes the input swarm metrics and dynamically recommends an adaptive DE search configuration. Specifically, it prescribes one or more candidate mutation strategies alongside their associated control parameters, namely the scaling factor ($F$) and crossover rate ($\mathit{CR}$). Driven by this LLM-recommended configuration, each target vector undergoes mutation and crossover operations to generate its corresponding trial vector, ultimately constructing the complete offspring population.

    \item \textbf{Population Selection and Closed-Loop Feedback}: 
    The generated trial vectors are evaluated and compared against their parent counterparts via a greedy selection mechanism, retaining superior individuals to update the population. The updated swarm state is subsequently routed back to the LLM, enabling real-time adjustment of strategy selection and parameter configurations based on observed evolutionary performance. This procedure iterates continuously until the termination criterion is satisfied, at which point the algorithm outputs the best solution found along with its objective value.

\end{itemize}

Through continuous inter-module feedback, LLMDE dynamically regulates its search behavior to seamlessly transition between exploration and exploitation phases, thereby elevating both its global optimization capability and environmental adaptability. To balance computational efficiency with adaptive search control, the LLM module is invoked periodically at an interval of $\Delta T_{\mathrm{LLM}}$ generations. During each invocation, the size and structure of the LLM’s output configuration set are explicitly constrained to ensure format consistency while preserving search stochasticity. Specifically, the number of prescribed parameter groups $K$ is dynamically determined by scaling the candidate operator pool size $\vert A \vert$ with a uniform random variable $r \in [0, 1]$, computed as $K = \lceil r \cdot \vert A \vert \rceil$. The pseudocode for the systematic implementation of LLMDE is provided in {Algorithm \ref{Algorithm1}}.

\begin{algorithm}[!t]

\caption{\rule[-5pt]{0pt}{16pt}The pseudocode of LLMDE}

\label{Algorithm1} 
\KwIn{Population size $N$, Problem dimension $D$, Maximum generation budget $T$, Candidate operator pool ${A} = \{op_1, op_2, \dots, op_{\vert {A} \vert}\}$, LLM invocation interval $\Delta T_{\mathrm{LLM}}$}
\KwOut{Optimum $\mathbf{x}_\mathrm{best}$ and $f(\mathbf{x}_\mathrm{best})$} 

\SetKwFunction{FMain}{LLMDE}
\SetKwProg{Fn}{Function}{:}{}

\Fn{\FMain{$N, D, T, A, \Delta T_{\mathrm{LLM}}$}}{

Initialize population $P=\{\mathbf{x}_1,\mathbf{x}_2,\ldots,\mathbf{x}_N\}$ using Eq.~\ref{Eq1} and evaluate individual fitness\;

$\mathbf{x}_\mathrm{best}\leftarrow
\arg\min_{\mathbf{x}_i\in P} f(\mathbf{x}_i)$\;

$t \leftarrow 1$\;

\While{$t < T$}
{
    Rank population $P$ according to fitness values\;

    \eIf{$t \pmod{\Delta T_{\mathrm{LLM}}} == 1$}{
        ${S}_t \leftarrow \textnormal{Calculate Swarm Status}(D,N,t/T,f_\mathrm{stagnation},P_\mathrm{std},f_\mathrm{best},f_\mathrm{avg},f_\mathrm{worst})$\;
        Sample $r \sim U(0, 1)$ and set output group size $K \leftarrow \lceil r \cdot |\mathcal{A}| \rceil$\;
        Query LLM to generate adaptive configuration set ${G}_t \leftarrow \textnormal{LLM}({S}_t,K, A)$\;
    }{
        Inherit search configuration set ${G}_{t} \leftarrow {G}_{t-1}$\;
    }

    \For{$i=1$ \KwTo $N$}
    {
        $(op_i,F_i,\mathit{CR}_i)\leftarrow \textnormal{Assigned Strategy}(i,{G}_t)$\;

        $\mathbf{v}_i\leftarrow 
        \textnormal{Mutation}(\mathbf{x}_i,op_i,F_i)$\;

        $\mathbf{u}_i\leftarrow \textnormal{Crossover}(\mathbf{x}_i, \mathbf{v}_i, \mathit{CR}_i) \textnormal{ using Eq.~(\ref{Eq3})}$\;

        $\mathbf{x}_i\leftarrow \textnormal{Selection}(\mathbf{u}_i, \mathbf{x}_i) \textnormal{ using Eq.~(\ref{Eq4})}$\;
    }

    $\mathbf{x}_\mathrm{best} \leftarrow \arg\min_{\mathbf{x}_i\in P} f(\mathbf{x}_i)$\;

    $t\leftarrow t+1$\;
}

\Return{$\mathbf{x}_\mathrm{best}$ \textnormal{and} $f(\mathbf{x}_\mathrm{best})$}\;

}
\end{algorithm}

\subsection{LLM Prompt Design}

The adaptive strategy configuration in LLMDE relies on encoding the real-time population state into a structured prompt, which is then fed into the LLM for contextual reasoning. Leveraging this state description, the LLM dynamically prescribes search operators alongside their parameter settings tailored for distinct population subgroups. Consequently, the synthesized strategies inherently adapt to the ongoing search dynamics and co-evolve with the optimization process. The structured prompt integrates three key components:

\begin{itemize}
    \item \textbf{Background Definition}: Establishes the fundamental context for strategy generation by specifying the LLM's role, primary task, and candidate decision domain. It comprises three core elements: First, the Role Definition designates the LLM as a ``Differential Evolution Strategy Architect," framing its reasoning from the perspective of an expert algorithm designer. Second, the Task clarifies its primary responsibility: evaluating real-time swarm status to dynamically allocate optimal search operators, scaling factors ($F$), and crossover rates ($\mathit{CR}$) for a balance between global exploration and local exploitation. Third, the Search Operator Pool explicitly bounds the candidate operator selection to standard DE variants (including DE/rand/2/bin, DE/rand/1/bin, DE/current-to-best/1/bin, DE/best/2/bin, and DE/best/1/bin), guaranteeing that all generated configurations remain structurally valid and algorithmically sound within the Differential Evolution framework. Together, these elements establish a clear task description and a well-defined decision space for subsequent strategy generation.

    \item \textbf{Operational Context}: Characterizes the real-time search status of the optimization process by capturing the population's evolutionary traits and performance metrics, thereby providing the necessary contextual feedback for LLM-driven adaptive decision-making. Specifically, it integrates key indicators that jointly reflect the search stage and population dynamics. Population diversity, measured by the standard deviation across the decision variable space, quantifies search space coverage and the algorithm's global exploration capacity. Stagnation counts the number of successive generations over which the best solution remains unimproved, serving as an indicator of search convergence stagnation. Additionally, fitness statistics (encompassing best, average, and worst values) provide insights into overall population convergence, performance distribution, and individual variation. By incorporating these contextual signals into the prompt, the LLM can effectively monitor optimization progress and generate adaptive strategies tailored to the current evolutionary state.

    \item \textbf{Assignment Directives}: Serves to regulate the strategy generation process by defining output structure and operational constraints to enable automated parsing by the optimization framework. Individuals are ranked by fitness, and the LLM divides the $N$ individuals into $K$ functional groups, assigning each group a search operator and its corresponding control parameters ($F$ and $\mathit{CR}$). To ensure syntactic validity, the response must be a standard JSON object comprising a brief reasoning field and a strategy list. Additionally, numerical precision is enforced by rounding $F$ and $\mathit{CR}$ to four decimal places and restricting the operator index to an integer within the predefined candidate pool.
\end{itemize}

\subsection{Computational complexity analysis}

The overall time complexity of LLMDE is dominated by population initialization, fitness evaluation, sorting, evolutionary operators, and periodic LLM invocations. The initialization takes $\mathcal{O}(N \cdot (D + f_{\mathrm{eval}}))$, where $N$ is the population size, $D$ is the problem dimension, and $f_{\mathrm{eval}}$ denotes the evaluation cost per solution. Across $T$ generations, the main loop requires $\mathcal{O}(T \cdot N \log N)$ for population ranking, $\mathcal{O}(T \cdot N \cdot (D + f_{\mathrm{eval}}))$ for mutation, crossover, and selection, and $\mathcal{O}(\frac{T}{\Delta T_{\mathrm{LLM}}} \cdot T_{\mathrm{LLM}})$ for LLM query overheads, where $\Delta T_{\mathrm{LLM}}$ is the invocation interval and $T_{\mathrm{LLM}}$ represents the computational cost per LLM call. Therefore, the time complexity of LLMDE is $\mathcal{O}\left( T \cdot N \cdot (D + f_{\mathrm{eval}} + \log N) + \frac{T}{\Delta T_{\mathrm{LLM}}} \cdot T_{\mathrm{LLM}} \right)$.

The memory complexity of LLMDE is determined by the local memory requirements during evolution. Because the LLM is queried through remote APIs, its model parameters and activation states incur no local storage overhead. Consequently, runtime memory consumption is primarily governed by maintaining the population and intermediate evolutionary arrays. Specifically, storing a population of size $N$ with dimension $D$ consumes $\mathcal{O}(N \cdot D)$ memory. In addition, sorting and ranking operations require temporary storage of $\mathcal{O}(N)$ to manage fitness evaluations and candidate indices. Thus, the overall memory complexity of LLMDE is $\mathcal{O}(N \cdot D)$.

\section{Experiments} 

In this section, we conduct extensive experiments on the CEC2022 benchmark suite to comprehensively evaluate the performance of the proposed LLMDE algorithm.

\subsection{Experimental Setups}

The CEC2022 benchmark suite \cite{Kumar2022}, summarized in Table~\ref{Tab2}, is employed for evaluation via the Opfunu library \cite{nguyen2020framework}. For each benchmark function, the problem dimension is set to $D = 20$, the population size is fixed at $N = 100$, and the maximum number of function evaluations (MaxFEs) is set to $1000 \times D$. To ensure statistical soundness, each algorithm is executed 30 times independently. To assess competitive performance, LLMDE is tested against five representative DE variants employing different mutation strategies: DE/rand/2/bin (DE\_S0), DE/rand/1/bin (DE\_S1), DE/current-to-best/1/bin (DE\_S2), DE/best/2/bin (DE\_S3), and DE/best/1/bin (DE\_S4). For all baseline variants, the scaling factor $F$ and crossover rate $CR$ are fixed at 0.5.

\begin{table}[!t]
\centering
\scriptsize
\caption{Benchmark functions in the CEC2022 test suite.}
\label{Tab2}
\begin{tabular*}{\textwidth}{@{\extracolsep{\fill}} c l l c @{}}
\toprule
No. & Type & Function & $f_{\min}$ \\
\midrule
F1  & Unimodal   & Shifted and full Rotated Zakharov function                    & 300  \\
F2  & Multimodal & Shifted and full Rotated Rosenbrock's function                & 400  \\
F3  & Multimodal & Shifted and full Rotated Expanded Schaffer's F6 function      & 600  \\
F4  & Multimodal & Shifted and full Rotated Non-Continuous Rastrigin's function  & 800  \\
F5  & Multimodal & Shifted and full Rotated Levy function                        & 900  \\
F6  & Hybrid     & Hybrid function 1 ($N=3$)                                     & 1800 \\
F7  & Hybrid     & Hybrid function 2 ($N=6$)                                     & 2000 \\
F8  & Hybrid     & Hybrid function 3 ($N=5$)                                     & 2200 \\
F9  & Composition& Composition function 1 ($N=5$)                                & 2300 \\
F10 & Composition& Composition function 2 ($N=4$)                                & 2400 \\
F11 & Composition& Composition function 3 ($N=5$)                                & 2600 \\
F12 & Composition& Composition function 4 ($N=6$)                                & 2700 \\
\bottomrule
\end{tabular*}
\end{table}

\subsection{LLM Model Selection}

To identify the optimal large language model (LLM) backbone for the proposed LLMDE algorithm, we evaluated four candidate models on the CEC2022 benchmark suite: DeepSeek-V4-Flash and DeepSeek-V4-Pro \cite{xu2026deepseek}, alongside GPT-5.6-Luna and GPT-5.6-Terra \cite{openai_models_documentation}. All LLMs were accessed via their official APIs with the default temperature parameter set to 1.0. Table~\ref{Tab3} details the configuration information of the large language models. All experiments shared identical experimental configurations and invocation interval ($\Delta T_{\mathrm{LLM}} = 10$) to ensure a fair comparison. The empirical results are summarized in Table~\ref{Tab4}, where ``Mean" and ``Std" denote the average and standard deviation of the final objective values over independent runs, respectively. For each benchmark function, the best performance is bolded with a light-yellow background. Additionally, the Wilcoxon signed-rank test \cite{derrac2011practical} at a 0.05 significance level was conducted to verify statistical significance, where ``$>$", ``$=$", and ``$<$" denote that DeepSeek-V4-Flash performs significantly better than, statistically similarly to, and significantly worse than the compared model, respectively. As shown in the results, DeepSeek-V4-Flash achieves superior optimization accuracy across the majority of benchmark functions and ranks first overall in the Friedman test \cite{derrac2011practical} based on mean fitness values, demonstrating stronger optimization guidance under the current framework. Consequently, DeepSeek-V4-Flash is selected as the default configuration for subsequent experiments.

\begin{table}[htbp]
    \centering
    \caption{Configurations of Large Language Models.}
    \label{Tab3}
    \scriptsize
    \begin{tabular}{lllll}
    \toprule
    \textbf{Algorithm} & \textbf{Model} & \textbf{Developer} & \textbf{Access Period} & \textbf{API} \\
    \midrule
    \multirow{4}{*}{LLMDE} 
     & DeepSeek-V4-Flash & DeepSeek & Aug--Sep 2026 & \texttt{deepseek-v4-flash} \\
     & DeepSeek-V4-Pro   & DeepSeek & Aug--Sep 2026 & \texttt{deepseek-v4-pro} \\
     & GPT-5.6-Luna      & OpenAI   & Aug--Sep 2026 & \texttt{gpt-5.6-luna} \\
     & GPT-5.6-Terra     & OpenAI   & Aug--Sep 2026 & \texttt{gpt-5.6-terra} \\
    \bottomrule
\end{tabular}
\end{table}

\begin{table}[htb]
    \centering
    \caption{Comparative results of LLMDE equipped with different LLMs on CEC2022.}
    \renewcommand{\arraystretch}{0.9}  
    \scriptsize
    \setlength{\tabcolsep}{3pt} 
    \label{Tab4}
    \hspace*{-0.5cm}
    \begin{tabular}{ccccccccc}
    \toprule
    \multicolumn{1}{l}{\textbf{Fun.}} &       & \textbf{LLMDE-V4-Flash} & \textbf{LLMDE-V4-Pro} &       & \textbf{LLMDE-3.5-Luna} &       & \textbf{LLMDE-3.5-Tera} &  \\
    \midrule
    \multicolumn{1}{l}{\multirow{2}[0]{*}{F1}} & \multicolumn{1}{l}{Mean} & \cellcolor[rgb]{ .992,  .996,  .808}\textbf{6.394E+02} & 1.562E+03 & \multicolumn{1}{l}{$>$} & 1.963E+03 & \multicolumn{1}{l}{$>$} & 8.773E+02 & \multicolumn{1}{l}{=} \\
          & \multicolumn{1}{l}{Std} & \cellcolor[rgb]{ .992,  .996,  .808}\textbf{5.530E+02} & 1.303E+03 &       & 9.226E+02 &       & 6.706E+02 &  \\
    \multicolumn{1}{l}{\multirow{2}[0]{*}{F2}} & \multicolumn{1}{l}{Mean} & 4.492E+02 & \cellcolor[rgb]{ .992,  .996,  .808}\textbf{4.487E+02} & \multicolumn{1}{l}{=} & 4.510E+02 & \multicolumn{1}{l}{=} & \cellcolor[rgb]{ .992,  .996,  .808}\textbf{4.487E+02} & \multicolumn{1}{l}{=} \\
    \multicolumn{1}{l}{} & \multicolumn{1}{l}{Std} & 5.209E+00 & 8.341E+00 &       & 1.409E+01 &       & \cellcolor[rgb]{ .992,  .996,  .808}\textbf{1.408E+00} &  \\
    \multicolumn{1}{l}{\multirow{2}[0]{*}{F3}} & \multicolumn{1}{l}{Mean} & \cellcolor[rgb]{ .992,  .996,  .808}\textbf{6.000E+02} & \cellcolor[rgb]{ .992,  .996,  .808}\textbf{6.000E+02} & \multicolumn{1}{l}{$>$} & \cellcolor[rgb]{ .992,  .996,  .808}\textbf{6.000E+02} & \multicolumn{1}{l}{$>$} & \cellcolor[rgb]{ .992,  .996,  .808}\textbf{6.000E+02} & \multicolumn{1}{l}{$>$} \\
    \multicolumn{1}{l}{} & \multicolumn{1}{l}{Std} & \cellcolor[rgb]{ .992,  .996,  .808}\textbf{6.251E-07} & 1.497E-04 &       & 3.192E-05 &       & 5.497E-06 &  \\
    \multicolumn{1}{l}{\multirow{2}[0]{*}{F4}} & \multicolumn{1}{l}{Mean} & 8.039E+02 & \cellcolor[rgb]{ .992,  .996,  .808}\textbf{8.031E+02} & \multicolumn{1}{l}{$<$} & 8.039E+02 & \multicolumn{1}{l}{=} & 8.039E+02 & \multicolumn{1}{l}{=} \\
    \multicolumn{1}{l}{} & \multicolumn{1}{l}{Std} & \cellcolor[rgb]{ .992,  .996,  .808}\textbf{3.531E-01} & 7.842E-01 &       & 3.792E-01 &       & 4.808E-01 &  \\
    \multicolumn{1}{l}{\multirow{2}[0]{*}{F5}} & \multicolumn{1}{l}{Mean} & \cellcolor[rgb]{ .992,  .996,  .808}\textbf{9.000E+02} & 9.006E+02 & \multicolumn{1}{l}{$>$} & 9.003E+02 & \multicolumn{1}{l}{$>$} & 9.001E+02 & \multicolumn{1}{l}{=} \\
    \multicolumn{1}{l}{} & \multicolumn{1}{l}{Std} & \cellcolor[rgb]{ .992,  .996,  .808}\textbf{6.247E-02} & 1.032E+00 &       & 2.316E-01 &       & 1.448E-01 &  \\
    \multicolumn{1}{l}{\multirow{2}[0]{*}{F6}} & \multicolumn{1}{l}{Mean} & \cellcolor[rgb]{ .992,  .996,  .808}\textbf{6.022E+06} & 9.457E+06 & \multicolumn{1}{l}{=} & 2.263E+07 & \multicolumn{1}{l}{$>$} & 1.825E+07 & \multicolumn{1}{l}{=} \\
    \multicolumn{1}{l}{} & \multicolumn{1}{l}{Std} & \cellcolor[rgb]{ .992,  .996,  .808}\textbf{8.410E+06} & 1.009E+07 &       & 1.785E+07 &       & 1.198E+07 &  \\
    \multicolumn{1}{l}{\multirow{2}[0]{*}{F7}} & \multicolumn{1}{l}{Mean} & \cellcolor[rgb]{ .992,  .996,  .808}\textbf{2.152E+03} & 2.205E+03 & \multicolumn{1}{l}{=} & 2.321E+03 & \multicolumn{1}{l}{$>$} & 2.248E+03 & \multicolumn{1}{l}{=} \\
    \multicolumn{1}{l}{} & \multicolumn{1}{l}{Std} & 9.295E+01 & \cellcolor[rgb]{ .992,  .996,  .808}\textbf{7.983E+01} &       & 1.605E+02 &       & 1.289E+02 &  \\
    \multicolumn{1}{l}{\multirow{2}[0]{*}{F8}} & \multicolumn{1}{l}{Mean} & \cellcolor[rgb]{ .992,  .996,  .808}\textbf{3.155E+03} & 4.541E+03 & \multicolumn{1}{l}{=} & 8.653E+03 & \multicolumn{1}{l}{$>$} & 6.716E+03 & \multicolumn{1}{l}{$>$} \\
    \multicolumn{1}{l}{} & \multicolumn{1}{l}{Std} & \cellcolor[rgb]{ .992,  .996,  .808}\textbf{1.131E+03} & 2.434E+03 &       & 1.055E+04 &       & 3.323E+03 &  \\
    \multicolumn{1}{l}{\multirow{2}[0]{*}{F9}} & \multicolumn{1}{l}{Mean} & \cellcolor[rgb]{ .992,  .996,  .808}\textbf{2.637E+03} & 2.640E+03 & \multicolumn{1}{l}{=} & 2.642E+03 & \multicolumn{1}{l}{$>$} & 2.637E+03 & \multicolumn{1}{l}{=} \\
    \multicolumn{1}{l}{} & \multicolumn{1}{l}{Std} & 1.184E+00 & 5.283E+00 &       & 5.401E+00 &       & \cellcolor[rgb]{ .992,  .996,  .808}\textbf{1.162E+00} &  \\
    \multicolumn{1}{l}{\multirow{2}[0]{*}{F10}} & \multicolumn{1}{l}{Mean} & 3.190E+03 & 2.914E+03 & \multicolumn{1}{l}{=} & 3.362E+03 & \multicolumn{1}{l}{=} & \cellcolor[rgb]{ .992,  .996,  .808}\textbf{2.912E+03} & \multicolumn{1}{l}{=} \\
    \multicolumn{1}{l}{} & \multicolumn{1}{l}{Std} & 1.157E+03 & \cellcolor[rgb]{ .992,  .996,  .808}\textbf{6.610E+02} &       & 1.441E+03 &       & 7.021E+02 &  \\
    \multicolumn{1}{l}{\multirow{2}[0]{*}{F11}} & \multicolumn{1}{l}{Mean} & \cellcolor[rgb]{ .992,  .996,  .808}\textbf{2.601E+03} & 2.674E+03 & \multicolumn{1}{l}{$>$} & 2.621E+03 & \multicolumn{1}{l}{=} & 2.602E+03 & \multicolumn{1}{l}{=} \\
    \multicolumn{1}{l}{} & \multicolumn{1}{l}{Std} & \cellcolor[rgb]{ .992,  .996,  .808}\textbf{1.580E+00} & 2.680E+02 &       & 5.821E+01 &       & 6.347E+00 &  \\
    \multicolumn{1}{l}{\multirow{2}[0]{*}{F12}} & \multicolumn{1}{l}{Mean} & \cellcolor[rgb]{ .992,  .996,  .808}\textbf{2.942E+03} & 2.944E+03 & \multicolumn{1}{l}{=} & 2.944E+03 & \multicolumn{1}{l}{=} & 2.944E+03 & \multicolumn{1}{l}{=} \\
    \multicolumn{1}{l}{} & \multicolumn{1}{l}{Std} & \cellcolor[rgb]{ .992,  .996,  .808}\textbf{3.290E+00} & 6.624E+00 &       & 8.939E+00 &       & 8.034E+00 &  \\
    \midrule
    \multicolumn{2}{c}{\textbf{$>/=/<$}} & \textbf{compared} & \textbf{ 4/7/1} &       & \textbf{ 7/5/0} &       & \textbf{ 2/10/0} &  \\
    \multicolumn{2}{c}{\textbf{Ranking}} & \textbf{1} & \textbf{3} &       & \textbf{4} &       & \textbf{2} &  \\
    \bottomrule
    \end{tabular}%
\end{table}%

\subsection{Sensitivity Analysis of $\Delta T_{\mathrm{LLM}}$}

Having selected the base LLM, we further investigated the impact of LLM call frequency on algorithm performance. The calling frequency governs the extent to which the LLM intervenes in the evolutionary search and directly dictates the additional computational overhead. Accordingly, we evaluated three invocation intervals: 5, 10, and 20 generations. To ensure a fair comparison, all other experimental parameters remained strictly identical across configurations. The comparative results are summarized in Table~\ref {Tab5}, where the three settings are assessed via mean values, standard deviations, Wilcoxon signed-rank tests, and overall rankings. As shown, $\Delta T_{\mathrm{LLM}} = 5$ demonstrates the most competitive performance, consistently yielding superior mean quality with lower variance across the majority of benchmark functions, while securing the top overall rank. These findings suggest that a shorter invocation interval allows the LLM to capture the ongoing evolutionary state more promptly and adapt the search dynamics according to up-to-date optimization feedback, thereby enhancing overall search efficiency. We therefore set the invocation interval $\Delta T_{\mathrm{LLM}}$ to 5 in subsequent experiments.

\begin{table}[htbp]
    \centering
    \caption{Comparative results of LLMDE under different invocation intervals ($\Delta T_{\mathrm{LLM}}$) on CEC2022.}
    \renewcommand{\arraystretch}{0.9}  
    \scriptsize
    \setlength{\tabcolsep}{3pt} 
    \label{Tab5}
    \hspace*{-0.5cm}
    \begin{tabular}{ccccccc}
    \toprule
    \multicolumn{1}{l}{\textbf{Fun.}} &       & \textbf{$\Delta T_{\mathrm{LLM}} = 10$} & \textbf{$\Delta T_{\mathrm{LLM}} = 5$} &       & \textbf{$\Delta T_{\mathrm{LLM}} = 20$} \\
    \midrule
    \multicolumn{1}{l}{\multirow{2}[0]{*}{F1}} & \multicolumn{1}{l}{Mean} & 6.394E+02 & \cellcolor[rgb]{ .992,  .996,  .808}\textbf{3.600E+02} & =     & 8.714E+02 & = \\
          & \multicolumn{1}{l}{Std} & 5.530E+02 & \cellcolor[rgb]{ .992,  .996,  .808}\textbf{7.463E+01} &       & 1.610E+03 &  \\
    \multicolumn{1}{l}{\multirow{2}[0]{*}{F2}} & \multicolumn{1}{l}{Mean} & 4.492E+02 & 4.495E+02 & =     & \cellcolor[rgb]{ .992,  .996,  .808}\textbf{4.490E+02} & = \\
    \multicolumn{1}{l}{} & \multicolumn{1}{l}{Std} & 5.209E+00 & \cellcolor[rgb]{ .992,  .996,  .808}\textbf{4.411E+00} &       & 1.020E+01 &  \\
    \multicolumn{1}{l}{\multirow{2}[0]{*}{F3}} & \multicolumn{1}{l}{Mean} & \cellcolor[rgb]{ .992,  .996,  .808}\textbf{6.000E+02} & \cellcolor[rgb]{ .992,  .996,  .808}\textbf{6.000E+02} & =     & \cellcolor[rgb]{ .992,  .996,  .808}\textbf{6.000E+02} & $>$ \\
    \multicolumn{1}{l}{} & \multicolumn{1}{l}{Std} & 6.251E-07 & \cellcolor[rgb]{ .992,  .996,  .808}\textbf{3.511E-07} &       & 5.593E-06 &  \\
    \multicolumn{1}{l}{\multirow{2}[0]{*}{F4}} & \multicolumn{1}{l}{Mean} & 8.039E+02 & \cellcolor[rgb]{ .992,  .996,  .808}\textbf{8.036E+02} & =     & 8.037E+02 & = \\
    \multicolumn{1}{l}{} & \multicolumn{1}{l}{Std} & \cellcolor[rgb]{ .992,  .996,  .808}\textbf{3.531E-01} & 4.491E-01 &       & 5.624E-01 &  \\
    \multicolumn{1}{l}{\multirow{2}[0]{*}{F5}} & \multicolumn{1}{l}{Mean} & \cellcolor[rgb]{ .992,  .996,  .808}\textbf{9.000E+02} & \cellcolor[rgb]{ .992,  .996,  .808}\textbf{9.000E+02} & $<$     & 9.001E+02 & = \\
    \multicolumn{1}{l}{} & \multicolumn{1}{l}{Std} & 6.247E-02 & \cellcolor[rgb]{ .992,  .996,  .808}\textbf{3.033E-02} &       & 4.877E-01 &  \\
    \multicolumn{1}{l}{\multirow{2}[0]{*}{F6}} & \multicolumn{1}{l}{Mean} & 6.022E+06 & \cellcolor[rgb]{ .992,  .996,  .808}\textbf{2.782E+06} & =     & 6.221E+06 & = \\
    \multicolumn{1}{l}{} & \multicolumn{1}{l}{Std} & 8.410E+06 & \cellcolor[rgb]{ .992,  .996,  .808}\textbf{2.195E+06} &       & 1.081E+07 &  \\
    \multicolumn{1}{l}{\multirow{2}[0]{*}{F7}} & \multicolumn{1}{l}{Mean} & 2.152E+03 & \cellcolor[rgb]{ .992,  .996,  .808}\textbf{2.139E+03} & =     & 2.170E+03 & = \\
    \multicolumn{1}{l}{} & \multicolumn{1}{l}{Std} & 9.295E+01 & \cellcolor[rgb]{ .992,  .996,  .808}\textbf{5.641E+01} &       & 9.088E+01 &  \\
    \multicolumn{1}{l}{\multirow{2}[0]{*}{F8}} & \multicolumn{1}{l}{Mean} & 3.155E+03 & \cellcolor[rgb]{ .992,  .996,  .808}\textbf{3.050E+03} & =     & 4.144E+03 & = \\
    \multicolumn{1}{l}{} & \multicolumn{1}{l}{Std} & 1.131E+03 & \cellcolor[rgb]{ .992,  .996,  .808}\textbf{6.830E+02} &       & 2.063E+03 &  \\
    \multicolumn{1}{l}{\multirow{2}[0]{*}{F9}} & \multicolumn{1}{l}{Mean} & 2.637E+03 & \cellcolor[rgb]{ .992,  .996,  .808}\textbf{2.636E+03} & =     & 2.639E+03 & = \\
    \multicolumn{1}{l}{} & \multicolumn{1}{l}{Std} & 1.184E+00 & \cellcolor[rgb]{ .992,  .996,  .808}\textbf{9.011E-01} &       & 3.073E+00 &  \\
    \multicolumn{1}{l}{\multirow{2}[0]{*}{F10}} & \multicolumn{1}{l}{Mean} & 3.190E+03 & \cellcolor[rgb]{ .992,  .996,  .808}\textbf{2.800E+03} & =     & 3.323E+03 & = \\
    \multicolumn{1}{l}{} & \multicolumn{1}{l}{Std} & 1.157E+03 & \cellcolor[rgb]{ .992,  .996,  .808}\textbf{9.799E+01} &       & 1.277E+03 &  \\
    \multicolumn{1}{l}{\multirow{2}[0]{*}{F11}} & \multicolumn{1}{l}{Mean} & 2.601E+03 & \cellcolor[rgb]{ .992,  .996,  .808}\textbf{2.600E+03} & $<$     & 2.603E+03 & = \\
    \multicolumn{1}{l}{} & \multicolumn{1}{l}{Std} & 1.580E+00 & \cellcolor[rgb]{ .992,  .996,  .808}\textbf{9.667E-01} &       & 9.178E+00 &  \\
    \multicolumn{1}{l}{\multirow{2}[0]{*}{F12}} & \multicolumn{1}{l}{Mean} & 2.942E+03 & \cellcolor[rgb]{ .992,  .996,  .808}\textbf{2.940E+03} & =     & 2.943E+03 & = \\
    \multicolumn{1}{l}{} & \multicolumn{1}{l}{Std} & 3.290E+00 & \cellcolor[rgb]{ .992,  .996,  .808}\textbf{2.718E+00} &       & 7.810E+00 &  \\
    \midrule
    \multicolumn{2}{c}{\textbf{$>/=/<$}} & \textbf{compared} & \textbf{ 0/10/2} &       & \textbf{ 1/11/0} &  \\
    \multicolumn{2}{c}{\textbf{Ranking}} & \textbf{2} & \textbf{1} &       & \textbf{3} &  \\
    \bottomrule
    \end{tabular}
\end{table}

\subsection{Ablation Study}

To evaluate the efficacy of the proposed LLM-guided mechanism, particularly the individual contributions of population-state feedback and adaptive strategy selection, three algorithmic configurations were systematically compared. The proposed LLMDE invokes the LLM every five generations and provides population-state feedback (Operational Context), enabling the model to assess evolutionary dynamics and adaptively select appropriate mutation strategies. The ablation variant, LLMDE\_rOC, preserves the five-generation invocation schedule but removes the Operational Context, thereby isolating the role of population-state feedback in guiding strategy selection. Meanwhile, DE\_Srand employs a purely random strategy selection from the candidate pool every five generations without LLM involvement, serving as a baseline to verify whether the observed improvements stem from LLM reasoning rather than arbitrary switching.

The ablation results across the CEC2022 benchmark suite are summarized in Table~\ref {Tab6}. LLMDE achieves the best overall performance, securing the top rank. Pairwise Wilcoxon signed-rank tests show win/tie/loss records of $4/8/0$ against LLMDE\_rOC and $4/7/1$ against DE\_Srand, underscoring its statistical advantage over both baselines. In particular, the superiority over LLMDE\_rOC confirms that the operational context provides vital situational awareness for informed decision-making. Furthermore, outperforming DE\_Srand confirms that the performance gains stem from reasoned, state-dependent adaptation rather than the mere diversity benefits of periodic strategy alternation. Overall, these findings validate that both population-state feedback and LLM-guided adaptation are indispensable components of the framework, jointly bolstering the effectiveness of the proposed search paradigm.

\begin{table}[htbp]
    \centering
    \caption{Ablation study results of the proposed LLM-guided mechanism on CEC2022.}
    \renewcommand{\arraystretch}{0.9}  
    \scriptsize
    \setlength{\tabcolsep}{3pt} 
    \label{Tab6}
    \hspace*{-0.5cm}
    \begin{tabular}{ccccccc}
    \toprule
    \multicolumn{1}{l}{\textbf{Fun.}} &       & \textbf{LLMDE} & \textbf{LLMDE\_rOC} &       & \textbf{DE\_Srand} &  \\
    \midrule
    \multicolumn{1}{l}{\multirow{2}[0]{*}{F1}} & \multicolumn{1}{l}{Mean} & \cellcolor[rgb]{ .992,  .996,  .808}\textbf{3.600E+02} & 4.973E+02 & =     & 6.723E+02 & = \\
          & \multicolumn{1}{l}{Std} & \cellcolor[rgb]{ .992,  .996,  .808}\textbf{7.463E+01} & 5.371E+02 &       & 3.674E+02 &  \\
    \multicolumn{1}{l}{\multirow{2}[0]{*}{F2}} & \multicolumn{1}{l}{Mean} & 4.495E+02 & \cellcolor[rgb]{ .992,  .996,  .808}\textbf{4.466E+02} & =     & 4.530E+02 & = \\
    \multicolumn{1}{l}{} & \multicolumn{1}{l}{Std} & \cellcolor[rgb]{ .992,  .996,  .808}\textbf{4.411E+00} & 8.968E+00 &       & 1.231E+01 &  \\
    \multicolumn{1}{l}{\multirow{2}[0]{*}{F3}} & \multicolumn{1}{l}{Mean} & \cellcolor[rgb]{ .992,  .996,  .808}\textbf{6.000E+02} & \cellcolor[rgb]{ .992,  .996,  .808}\textbf{6.000E+02} & =     & \cellcolor[rgb]{ .992,  .996,  .808}\textbf{6.000E+02} & $<$ \\
    \multicolumn{1}{l}{} & \multicolumn{1}{l}{Std} & 3.511E-07 & 5.940E-06 &       & \cellcolor[rgb]{ .992,  .996,  .808}\textbf{1.229E-08} &  \\
    \multicolumn{1}{l}{\multirow{2}[0]{*}{F4}} & \multicolumn{1}{l}{Mean} & \cellcolor[rgb]{ .992,  .996,  .808}\textbf{8.036E+02} & 8.037E+02 & =     & 8.037E+02 & = \\
    \multicolumn{1}{l}{} & \multicolumn{1}{l}{Std} & 4.491E-01 & 5.669E-01 &       & \cellcolor[rgb]{ .992,  .996,  .808}\textbf{3.634E-01} &  \\
    \multicolumn{1}{l}{\multirow{2}[0]{*}{F5}} & \multicolumn{1}{l}{Mean} & \cellcolor[rgb]{ .992,  .996,  .808}\textbf{9.000E+02} & 9.001E+02 & $>$     & \cellcolor[rgb]{ .992,  .996,  .808}\textbf{9.000E+02} & = \\
    \multicolumn{1}{l}{} & \multicolumn{1}{l}{Std} & \cellcolor[rgb]{ .992,  .996,  .808}\textbf{3.033E-02} & 1.403E-01 &       & 9.837E-02 &  \\
    \multicolumn{1}{l}{\multirow{2}[0]{*}{F6}} & \multicolumn{1}{l}{Mean} & 2.782E+06 & \cellcolor[rgb]{ .992,  .996,  .808}\textbf{9.905E+05} & =     & 1.930E+07 & $>$ \\
    \multicolumn{1}{l}{} & \multicolumn{1}{l}{Std} & 2.195E+06 & \cellcolor[rgb]{ .992,  .996,  .808}\textbf{1.430E+06} &       & 1.448E+07 &  \\
    \multicolumn{1}{l}{\multirow{2}[0]{*}{F7}} & \multicolumn{1}{l}{Mean} & 2.139E+03 & \cellcolor[rgb]{ .992,  .996,  .808}\textbf{2.134E+03} & =     & 2.137E+03 & = \\
    \multicolumn{1}{l}{} & \multicolumn{1}{l}{Std} & \cellcolor[rgb]{ .992,  .996,  .808}\textbf{5.641E+01} & 6.411E+01 &       & 6.175E+01 &  \\
    \multicolumn{1}{l}{\multirow{2}[0]{*}{F8}} & \multicolumn{1}{l}{Mean} & 3.050E+03 & \cellcolor[rgb]{ .992,  .996,  .808}\textbf{2.968E+03} & =     & 8.730E+03 & $>$ \\
    \multicolumn{1}{l}{} & \multicolumn{1}{l}{Std} & \cellcolor[rgb]{ .992,  .996,  .808}\textbf{6.830E+02} & 9.355E+02 &       & 9.441E+03 &  \\
    \multicolumn{1}{l}{\multirow{2}[0]{*}{F9}} & \multicolumn{1}{l}{Mean} & \cellcolor[rgb]{ .992,  .996,  .808}\textbf{2.636E+03} & 2.637E+03 & =     & 2.639E+03 & $>$ \\
    \multicolumn{1}{l}{} & \multicolumn{1}{l}{Std} & \cellcolor[rgb]{ .992,  .996,  .808}\textbf{9.011E-01} & 1.312E+00 &       & 2.830E+00 &  \\
    \multicolumn{1}{l}{\multirow{2}[0]{*}{F10}} & \multicolumn{1}{l}{Mean} & \cellcolor[rgb]{ .992,  .996,  .808}\textbf{2.800E+03} & 3.294E+03 & $>$     & 3.073E+03 & = \\
    \multicolumn{1}{l}{} & \multicolumn{1}{l}{Std} & \cellcolor[rgb]{ .992,  .996,  .808}\textbf{9.799E+01} & 1.258E+03 &       & 9.304E+02 &  \\
    \multicolumn{1}{l}{\multirow{2}[0]{*}{F11}} & \multicolumn{1}{l}{Mean} & \cellcolor[rgb]{ .992,  .996,  .808}\textbf{2.600E+03} & 2.654E+03 & $>$     & 2.607E+03 & = \\
    \multicolumn{1}{l}{} & \multicolumn{1}{l}{Std} & \cellcolor[rgb]{ .992,  .996,  .808}\textbf{9.667E-01} & 2.536E+02 &       & 2.536E+01 &  \\
    \multicolumn{1}{l}{\multirow{2}[0]{*}{F12}} & \multicolumn{1}{l}{Mean} & \cellcolor[rgb]{ .992,  .996,  .808}\textbf{2.940E+03} & 2.947E+03 & $>$     & 2.944E+03 & $>$ \\
    \multicolumn{1}{l}{} & \multicolumn{1}{l}{Std} & \cellcolor[rgb]{ .992,  .996,  .808}\textbf{2.718E+00} & 1.139E+01 &       & 9.736E+00 &  \\
    \midrule
    \multicolumn{2}{c}{\textbf{$>/=/<$}} & \textbf{compared} & \textbf{4/8/0} &       & \textbf{ 4/7/1} &  \\
    \multicolumn{2}{c}{\textbf{Ranking}} & \textbf{1} & \textbf{2} &       & \textbf{3} &  \\
    \bottomrule
    \end{tabular}
\end{table}

\subsection{Comparison with Different DE Mutation Strategies}

LLMDE was evaluated against five classical DE mutation strategies: DE/rand/2/bin (DE\_S0), DE/rand/1/bin (DE\_S1), DE/current-to-best/1/bin (DE\_S2), DE/best/2/bin (DE\_S3), and DE/best/1/bin (DE\_S4). As summarized in Table~\ref{Tab7}, the proposed LLMDE algorithm achieves win/tie/loss counts of $12/0/0$, $10/2/0$, $3/6/3$, $8/4/0$, and $4/4/4$ against DE\_S0 through DE\_S4, respectively, securing the overall first rank across the benchmark suite. Among the baselines, DE\_S2 exhibits the strongest overall competitiveness and claims second place across the benchmark suite. Although DE\_S4 achieves superior mean performance on several specific landscapes (e.g., F1, F4, F6, and F7) via greedy exploitation, it suffers from notable performance degradation on more complex multimodal functions. These outcomes closely reflect the inherent exploration–exploitation trade-offs of the individual operators: while DE\_S0 and DE\_S1 rely primarily on stochastic exploration, DE\_S3 and DE\_S4 favor aggressive local exploitation at the expense of rapid diversity loss, leading to a higher risk of premature convergence. In contrast, DE\_S2 maintains a more balanced search trajectory. By dynamically coordinating these complementary strategies according to real-time search feedback, LLMDE leverages the exploratory and exploitative strengths of each operator, thereby achieving superior overall optimization performance.

\begin{table}[htbp]
    \centering
    \caption{Comparison results of LLMDE with different DE mutation strategies on CEC2022.}
    \renewcommand{\arraystretch}{0.9}  
    \scriptsize
    \setlength{\tabcolsep}{3pt} 
    \label{Tab7}
    \hspace*{-0.5cm}
    \begin{tabular}{ccccccccccccc}
    \toprule
    \multicolumn{1}{l}{\textbf{Fun.}} &       & \textbf{LLMDE} & \textbf{DE\_S0} &       & \textbf{DE\_S1} &       & \textbf{DE\_S2} &       & \textbf{DE\_S3} &       & \textbf{DE\_S4} &  \\
    \midrule
    \multicolumn{1}{l}{\multirow{2}[0]{*}{F1}} & \multicolumn{1}{l}{Mean} & 3.600E+02 & 1.933E+04 & $>$     & 9.064E+03 & $>$      & \cellcolor[rgb]{ .992,  .996,  .808}\textbf{3.000E+02} & $<$      & 2.817E+03 & $>$      & \cellcolor[rgb]{ .992,  .996,  .808}\textbf{3.000E+02} & $<$  \\
          & \multicolumn{1}{l}{Std} & 7.463E+01 & 3.733E+03 &       & 1.691E+03 &       & 2.742E-04 &       & 7.098E+02 &       & \cellcolor[rgb]{ .992,  .996,  .808}\textbf{1.768E-04} &  \\
    \multicolumn{1}{l}{\multirow{2}[0]{*}{F2}} & \multicolumn{1}{l}{Mean} & \cellcolor[rgb]{ .992,  .996,  .808}\textbf{4.495E+02} & 5.024E+02 & $>$      & 4.504E+02 & $>$      & 4.565E+02 & $>$      & 4.511E+02 & =     & 4.518E+02 & = \\
    \multicolumn{1}{l}{} & \multicolumn{1}{l}{Std} & 4.411E+00 & 1.196E+01 &       & \cellcolor[rgb]{ .992,  .996,  .808}\textbf{4.579E-01} &       & 1.016E+01 &       & 9.657E+00 &       & 1.375E+01 &  \\
    \multicolumn{1}{l}{\multirow{2}[0]{*}{F3}} & \multicolumn{1}{l}{Mean} & \cellcolor[rgb]{ .992,  .996,  .808}\textbf{6.000E+02} & \cellcolor[rgb]{ .992,  .996,  .808}\textbf{6.000E+02} & $>$      & \cellcolor[rgb]{ .992,  .996,  .808}\textbf{6.000E+02} & $>$      & \cellcolor[rgb]{ .992,  .996,  .808}\textbf{6.000E+02} & $<$      & \cellcolor[rgb]{ .992,  .996,  .808}\textbf{6.000E+02} & =     & \cellcolor[rgb]{ .992,  .996,  .808}\textbf{6.000E+02} & $<$  \\
    \multicolumn{1}{l}{} & \multicolumn{1}{l}{Std} & 3.511E-07 & 1.078E-03 &       & 5.002E-06 &       & \cellcolor[rgb]{ .992,  .996,  .808}\textbf{4.151E-14} &       & 1.957E-07 &       & 8.039E-14 &  \\
    \multicolumn{1}{l}{\multirow{2}[0]{*}{F4}} & \multicolumn{1}{l}{Mean} & 8.036E+02 & 8.040E+02 & $>$      & 8.038E+02 & =     & 8.033E+02 & =     & 8.038E+02 & =     & \cellcolor[rgb]{ .992,  .996,  .808}\textbf{8.030E+02} & $<$  \\
    \multicolumn{1}{l}{} & \multicolumn{1}{l}{Std} & 4.491E-01 & \cellcolor[rgb]{ .992,  .996,  .808}\textbf{3.434E-01} &       & 4.060E-01 &       & 4.866E-01 &       & 4.617E-01 &       & 1.031E+00 &  \\
    \multicolumn{1}{l}{\multirow{2}[0]{*}{F5}} & \multicolumn{1}{l}{Mean} & \cellcolor[rgb]{ .992,  .996,  .808}\textbf{9.000E+02} & 9.042E+02 & $>$      & 9.004E+02 & $>$      & \cellcolor[rgb]{ .992,  .996,  .808}\textbf{9.000E+02} & =     & 9.001E+02 & $>$      & 9.007E+02 & $>$  \\
    \multicolumn{1}{l}{} & \multicolumn{1}{l}{Std} & \cellcolor[rgb]{ .992,  .996,  .808}\textbf{3.033E-02} & 7.326E-01 &       & 9.033E-02 &       & 1.005E-01 &       & 1.850E-01 &       & 9.302E-01 &  \\
    \multicolumn{1}{l}{\multirow{2}[0]{*}{F6}} & \multicolumn{1}{l}{Mean} & 2.782E+06 & 2.500E+08 & $>$      & 1.385E+08 & $>$      & 3.557E+06 & =     & 4.656E+07 & $>$      & \cellcolor[rgb]{ .992,  .996,  .808}\textbf{1.484E+06} & = \\
    \multicolumn{1}{l}{} & \multicolumn{1}{l}{Std} & 2.195E+06 & 9.806E+07 &       & 5.775E+07 &       & 2.083E+06 &       & 2.507E+07 &       & \cellcolor[rgb]{ .992,  .996,  .808}\textbf{1.685E+06} &  \\
    \multicolumn{1}{l}{\multirow{2}[0]{*}{F7}} & \multicolumn{1}{l}{Mean} & 2.139E+03 & 2.732E+03 & $>$      & 2.512E+03 & $>$      & 2.072E+03 & $<$      & 2.270E+03 & $>$      & \cellcolor[rgb]{ .992,  .996,  .808}\textbf{2.068E+03} & $<$  \\
    \multicolumn{1}{l}{} & \multicolumn{1}{l}{Std} & 5.641E+01 & 1.934E+02 &       & 1.277E+02 &       & \cellcolor[rgb]{ .992,  .996,  .808}\textbf{2.219E+01} &       & 8.781E+01 &       & 2.409E+01 &  \\
    \multicolumn{1}{l}{\multirow{2}[0]{*}{F8}} & \multicolumn{1}{l}{Mean} & \cellcolor[rgb]{ .992,  .996,  .808}\textbf{3.050E+03} & 8.097E+05 & $>$      & 2.160E+05 & $>$      & 3.770E+03 & =     & 9.362E+03 & $>$      & 4.150E+03 & $>$  \\
    \multicolumn{1}{l}{} & \multicolumn{1}{l}{Std} & \cellcolor[rgb]{ .992,  .996,  .808}\textbf{6.830E+02} & 1.176E+06 &       & 3.683E+05 &       & 1.063E+03 &       & 4.972E+03 &       & 1.559E+03 &  \\
    \multicolumn{1}{l}{\multirow{2}[0]{*}{F9}} & \multicolumn{1}{l}{Mean} & \cellcolor[rgb]{ .992,  .996,  .808}\textbf{2.636E+03} & 2.719E+03 & $>$      & 2.675E+03 & $>$      & 2.639E+03 & $>$      & 2.643E+03 & $>$      & \cellcolor[rgb]{ .992,  .996,  .808}\textbf{2.636E+03} & = \\
    \multicolumn{1}{l}{} & \multicolumn{1}{l}{Std} & \cellcolor[rgb]{ .992,  .996,  .808}\textbf{9.011E-01} & 2.275E+01 &       & 1.140E+01 &       & 3.106E+00 &       & 5.003E+00 &       & 1.240E+00 &  \\
    \multicolumn{1}{l}{\multirow{2}[0]{*}{F10}} & \multicolumn{1}{l}{Mean} & \cellcolor[rgb]{ .992,  .996,  .808}\textbf{2.800E+03} & 3.403E+03 & $>$      & 2.982E+03 & $>$      & 3.040E+03 & =     & 3.767E+03 & $>$      & 4.086E+03 & $>$  \\
    \multicolumn{1}{l}{} & \multicolumn{1}{l}{Std} & \cellcolor[rgb]{ .992,  .996,  .808}\textbf{9.799E+01} & 9.891E+02 &       & 6.268E+02 &       & 9.072E+02 &       & 1.655E+03 &       & 1.770E+03 &  \\
    \multicolumn{1}{l}{\multirow{2}[0]{*}{F11}} & \multicolumn{1}{l}{Mean} & \cellcolor[rgb]{ .992,  .996,  .808}\textbf{2.600E+03} & 2.650E+03 & $>$      & 2.607E+03 & $>$      & \cellcolor[rgb]{ .992,  .996,  .808}\textbf{2.600E+03} & =     & 2.601E+03 & $>$      & 2.634E+03 & = \\
    \multicolumn{1}{l}{} & \multicolumn{1}{l}{Std} & 9.667E-01 & 1.339E+01 &       & 8.460E+00 &       & \cellcolor[rgb]{ .992,  .996,  .808}\textbf{8.169E-01} &       & 2.556E+00 &       & 1.329E+02 &  \\
    \multicolumn{1}{l}{\multirow{2}[0]{*}{F12}} & \multicolumn{1}{l}{Mean} & \cellcolor[rgb]{ .992,  .996,  .808}\textbf{2.940E+03} & 2.948E+03 & $>$      & \cellcolor[rgb]{ .992,  .996,  .808}\textbf{2.940E+03} & =     & 2.951E+03 & $>$      & 2.942E+03 & =     & 2.957E+03 & $>$  \\
    \multicolumn{1}{l}{} & \multicolumn{1}{l}{Std} & 2.718E+00 & 2.442E+00 &       & \cellcolor[rgb]{ .992,  .996,  .808}\textbf{1.701E+00} &       & 9.473E+00 &       & 6.706E+00 &       & 1.563E+01 &  \\
    \midrule
    \multicolumn{2}{c}{\textbf{$>/=/<$}} & \textbf{compared} & \textbf{ 12/0/0} &       & \textbf{ 10/2/0} &       & \textbf{ 3/6/3} &       & \textbf{8/4/0} &       & \textbf{ 4/4/4} &  \\
    \multicolumn{2}{c}{\textbf{Ranking}} & \textbf{1} & \textbf{6} &       & \textbf{5} &       & \textbf{2} &       & \textbf{4} &       & \textbf{3} &  \\
    \bottomrule
    \end{tabular}
\end{table}

\section{Application} 

This section validates the proposed LLMDE within a real-world stock investment scenario. We first filter promising candidate stocks from an initial investment pool to construct the target basket. Subsequently, portfolio optimization is conducted to evaluate the comparative performance of LLMDE against multiple baselines.

\subsection{Stock selection} 

We implement a systematic two-stage stock selection framework using firms' financial characteristics. Specifically, factor analysis is first applied to extract core financial dimensions, and $K$-means clustering is subsequently deployed to identify and screen superior-performing stocks.

\subsubsection{Sample Selection and Factor Analysis} 

The constituent stocks of the S\&P 500 index \cite{SP500Methodology2026} serve as the initial asset sample. Covering approximately 500 leading publicly traded U.S. firms across major economic sectors, the index offers superior market liquidity and high financial reporting transparency. To refine this initial pool, Return on Equity (ROE) is utilized as a screening criterion for profitability and capital efficiency, where filtering for firms with an ROE exceeding 18\% yields a target sample of 181 companies. For each firm, 13 representative financial metrics were collected to capture multi-dimensional corporate fundamentals. All financial data were retrieved from Yahoo Finance \cite{YahooFinance2026} as of March 11, 2026.

\begin{figure}[htbp]
    \centering
    \includegraphics[width=0.65\textwidth]{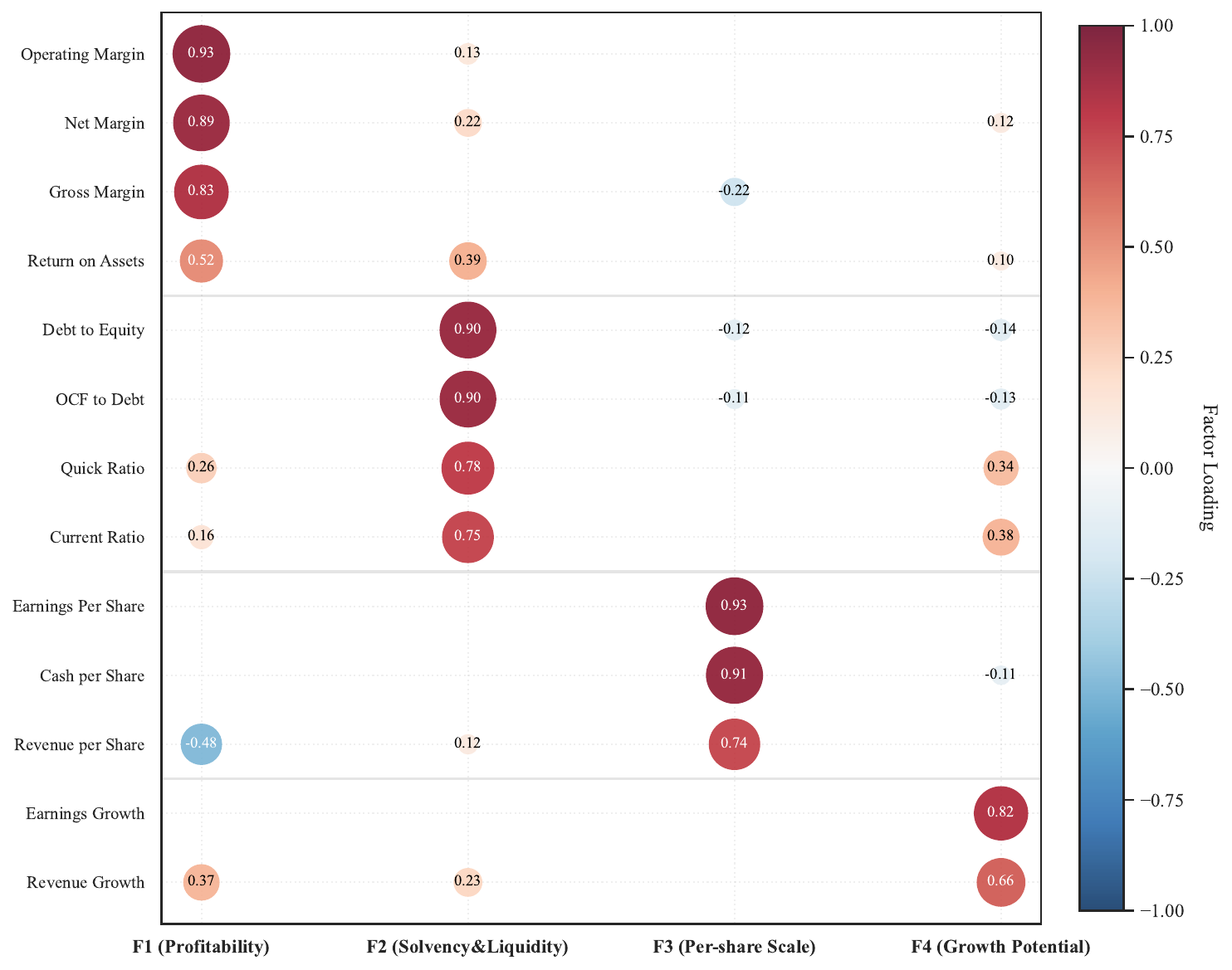} 
    \caption{Factor loadings of financial metrics.}
    \label{Fig2}  
\end{figure}

Prior to conducting factor analysis, directional alignment was performed on all collected raw financial data. The sampling adequacy and factorability of the metrics were verified using the Kaiser–Meyer–Olkin (KMO) measure and Bartlett's test of sphericity \cite{xi2024factor}. The resulting KMO statistic of 0.698 and a statistically significant Bartlett's test ($p < 0.001$) confirm that the correlation structure is well-suited for factor extraction. We subsequently employed factor analysis with Varimax orthogonal rotation to reduce dimensionality and yield more distinct, economically interpretable factor loadings\cite{kaiser1958varimax}. Four common factors were extracted, accounting for a cumulative variance of 76.7\% and capturing the vast majority of the original financial information. Figure~\ref{Fig2} illustrates the factor loadings of the financial metrics, where larger absolute values indicate a stronger correlation between the observed indicators and the latent factors. In accordance with the factor loading patterns, the economic interpretations of the four extracted dimensions are defined as follows:

\begin{itemize}[itemsep=1ex] 
    \item \textbf{F1 (Profitability):} Heavily loaded on Operating Margin, Net Margin, and Gross Margin, reflecting core operating efficiency and profit generation.
    
    \item \textbf{F2 (Solvency \& Liquidity):} Covers Debt-to-Equity and Liquidity-related metrics, reflecting the capital structure and asset liquidity.
    
    \item \textbf{F3 (Per-Share Scale):} Focuses on Earnings Per Share, Cash Per Share, and Revenue Per Share, representing the intrinsic value of individual shares.
    
    \item \textbf{F4 (Growth Potential):} Primarily driven by Earnings Growth and Revenue Growth, characterizing the firm's business expansion dynamics.
\end{itemize}

\subsubsection{Cluster-based Stock Screening} 

\begin{figure}[htbp]
    \centering    
    \includegraphics[width=0.65\textwidth]{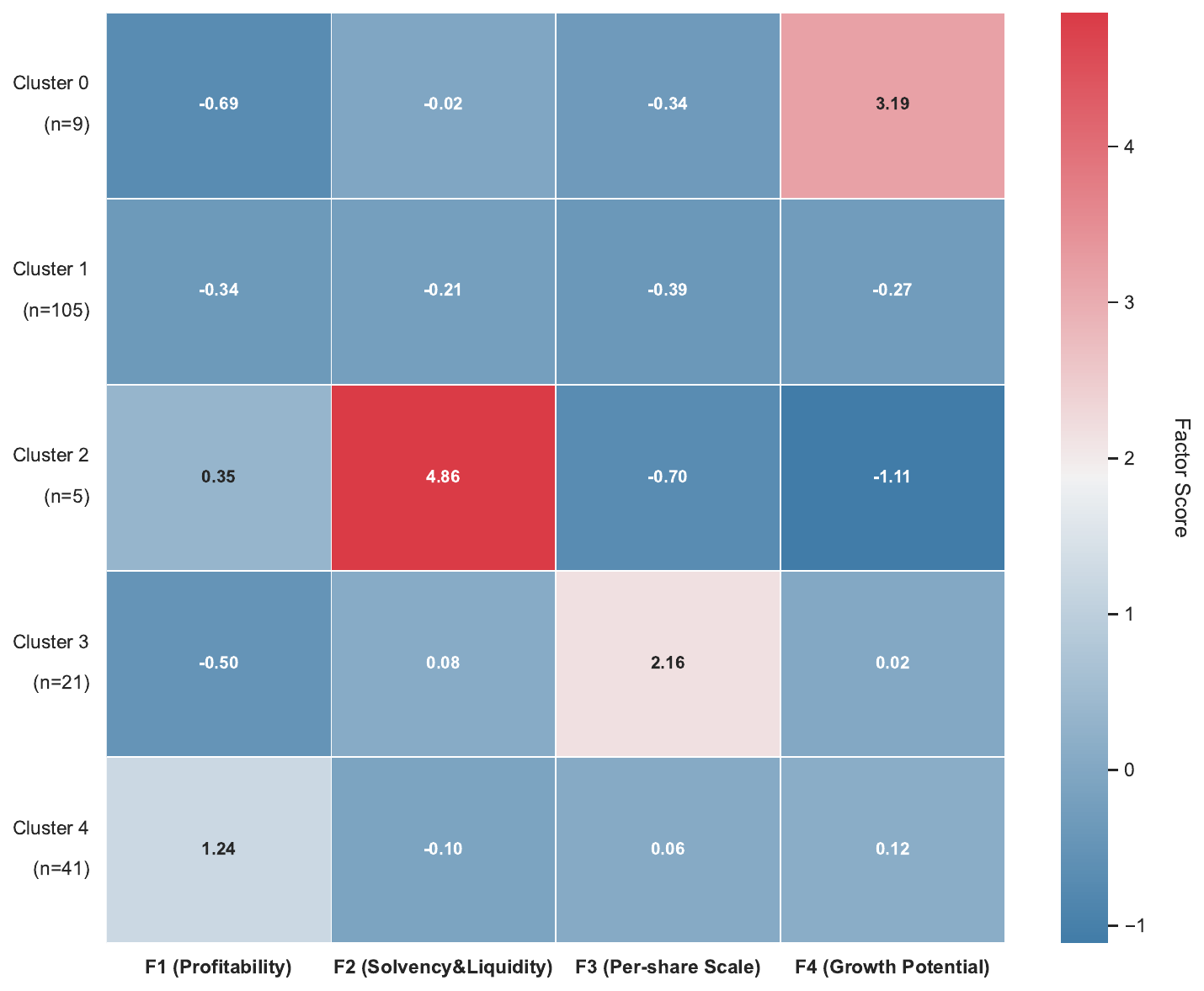}
    \caption{The Heatmap of asset clusters.}
    \label{Fig3}  
\end{figure}

Using the four extracted factor scores, firm-level feature vectors were constructed to partition the asset pool via the K-means clustering \cite{wu2022construction,chen2023credit}. The elbow method identified five as the optimal number of clusters \cite{thorndike1953belongs}. To characterize the distinct financial profiles of each cluster, Figure~\ref{Fig3} visualizes the average factor score distribution across the five groups (where n denotes cluster size) \cite{waskom2021seaborn}:

\begin{itemize}[itemsep=1.5ex] 
    \item \textbf{Cluster 0 (n = 9): Aggressive Growth} \\
    Firms exhibit low current profitability, while demonstrating exceptionally strong growth potential (F4 = 3.19). This pattern characterizes a typical high-growth, high-risk profile where future expansion is prioritized over immediate performance.

    \item \textbf{Cluster 1 (n = 105): Underperforming} \\
    This cluster underperforms across all four factors, revealing pervasive fundamental weakness and an absence of competitive advantage.

    \item \textbf{Cluster 2 (n = 5): Defensive} \\
    The financial performance of these firms indicates strong solvency and liquidity(F2 = 4.86), while maintaining a certain level of profitability, thereby reflecting strong financial stability and resilience against risks.

    \item \textbf{Cluster 3 (n = 21): High Per-Share Performance} \\
    These companies are characterized by outstanding per-share metrics (F3 = 2.16), indicating strong per-share earnings and cash-generation capabilities. 

    \item \textbf{Cluster 4 (n = 41): High-Profitability} \\
    The firms are chiefly distinguished by their outstanding profitability (F1 = 1.24). They show strong earnings generation, cost control, and efficient asset use, reflecting an overall robust and sustainable financial performance.
    
\end{itemize}

Following the cluster partitioning, Cluster 1 was excluded from the candidate universe due to its uniformly inferior performance across all dimensions. To screen high-quality assets from the remaining clusters, we computed a composite factor score defined as the weighted linear combination of the four factor scores, where the weights correspond to their respective variance contribution rates. This metric summarizes a firm's multi-dimensional fundamental strength across profitability, solvency, intrinsic value, and growth. Assets were first screened by imposing a minimum composite factor score threshold of 0.5. To ensure sectoral diversification, eligible firms were then selected across different industries, ultimately establishing a target investment basket of 12 constituent stocks. Historical daily adjusted closing prices for these assets were collected from January 1, 2025, to March 11, 2026, to calculate daily returns for subsequent empirical testing. Table~\ref{Tab8} summarizes the selected ticker symbols alongside their corresponding average daily returns.

\begin{table}[htbp] 
    \centering 
    \caption{Selected Stock Data} 
    \label{Tab8}
    \setlength{\tabcolsep}{4pt}
    \scriptsize 
    \renewcommand{\arraystretch}{0.9}  
    \begin{tabular}{lc} 
        \toprule 
       \textbf{Ticker} & \textbf{Mean Return}\\  
        \midrule 
        FIX   & 0.00405 \\
        GWW   & 0.00021 \\
        LLY   & 0.00090 \\
        META  & 0.00039 \\
        MNST  & 0.00131 \\
        MSFT  & -0.00010 \\
        NEM   & 0.00390 \\
        NVDA  & 0.00110 \\
        NVR   & -0.00072 \\
        TPL   & 0.00121 \\
        TROW  & -0.00062 \\
        URI   & 0.00038 \\
        \bottomrule 
    \end{tabular}
\end{table}

\subsection{Portfolio Optimization}

The portfolio optimization problem is formulated based on the CVaR model defined in Eq.~\ref{Eq8}, where CVaR serves as the risk metric. The optimization is subject to two fundamental constraints: a minimum target expected portfolio return of $5\%$ and a budget constraint ensuring that the portfolio weights sum to one. Both constraints are enforced via a penalty function, allowing the algorithm to minimize tail risk while satisfying feasibility requirements. The efficacy of CVaR depends heavily on the confidence level $\alpha$, as higher levels place greater weight on low-probability, high-impact tail losses. Accordingly, experiments were conducted across four representative confidence levels ($75\%$, $90\%$, $95\%$, and $99\%$) to thoroughly assess LLMDE under varying risk tolerances. To evaluate performance, LLMDE is compared with five DE variants and four representative metaheuristic algorithms: Genetic Algorithm (GA) \cite{holland1992adaptation}, Particle Swarm Optimization (PSO) \cite{kennedy1995particle}, Grey Wolf Optimizer (GWO) \cite{mirjalili2014grey}, and Sine Cosine Algorithm (SCA) \cite{mirjalili2016sca}. The experimental setup follows that of the CEC2022 experiments, with baseline parameters directly adopted from their original literature.

The optimization results across various CVaR confidence levels are summarized in Table~\ref{Tab9}. Overall, the proposed LLMDE demonstrates the most competitive performance among all evaluated algorithms. LLMDE secures the minimum CVaR at the $75\%$, $90\%$, and $95\%$ confidence levels, with PSO exhibiting a marginal advantage only at $99\%$. Furthermore, the optimal CVaR values naturally increase with higher confidence levels, which aligns well with the theoretical properties of CVaR, as higher confidence levels assign greater weight to extreme tail losses. These empirical findings confirm that LLMDE provides an effective framework for solving CVaR portfolio optimization under diverse risk preferences.

\begin{figure*}[h]
    \centering
    \includegraphics[width=0.48\textwidth]{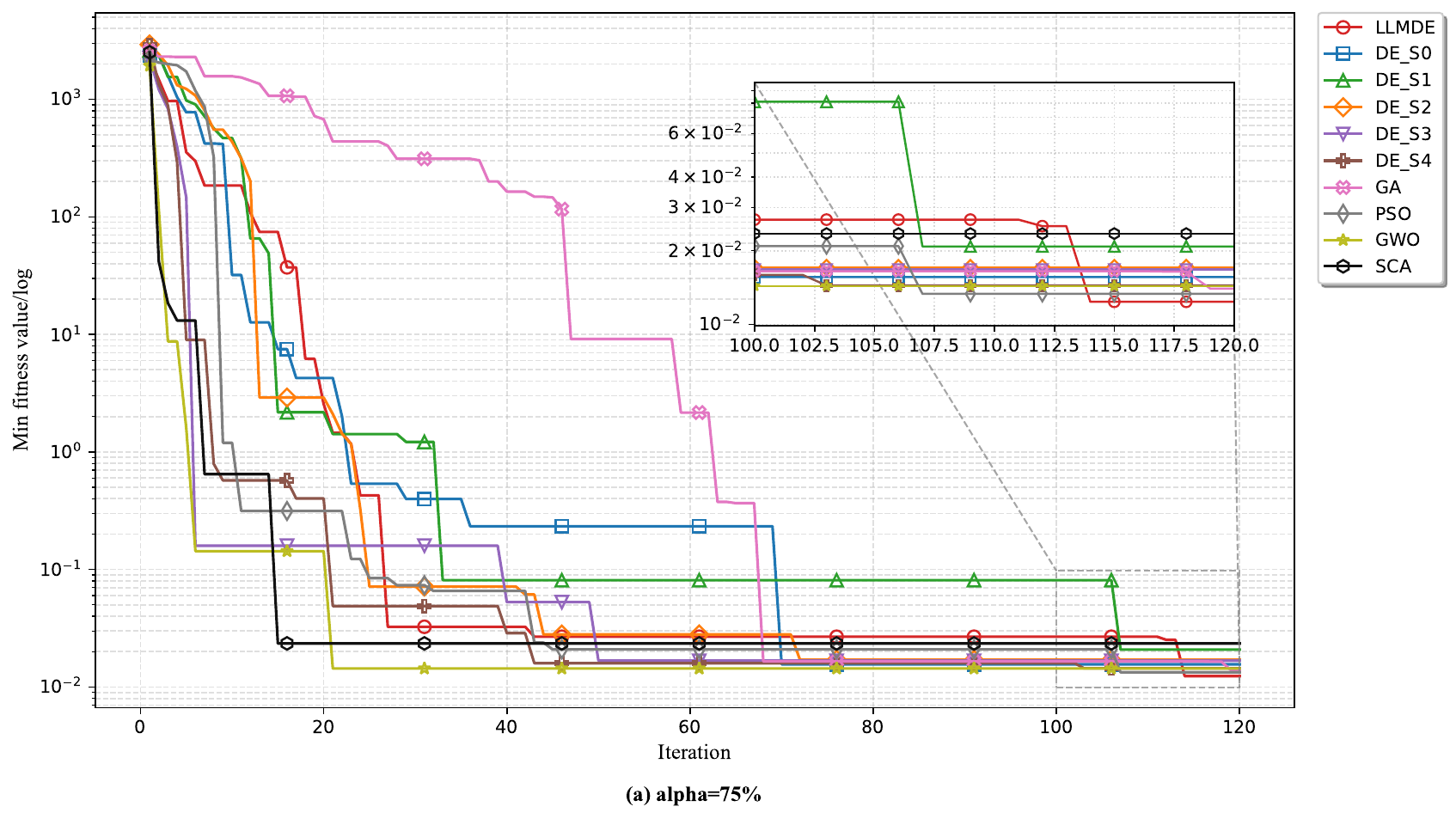}\hfill
    \includegraphics[width=0.48\textwidth]{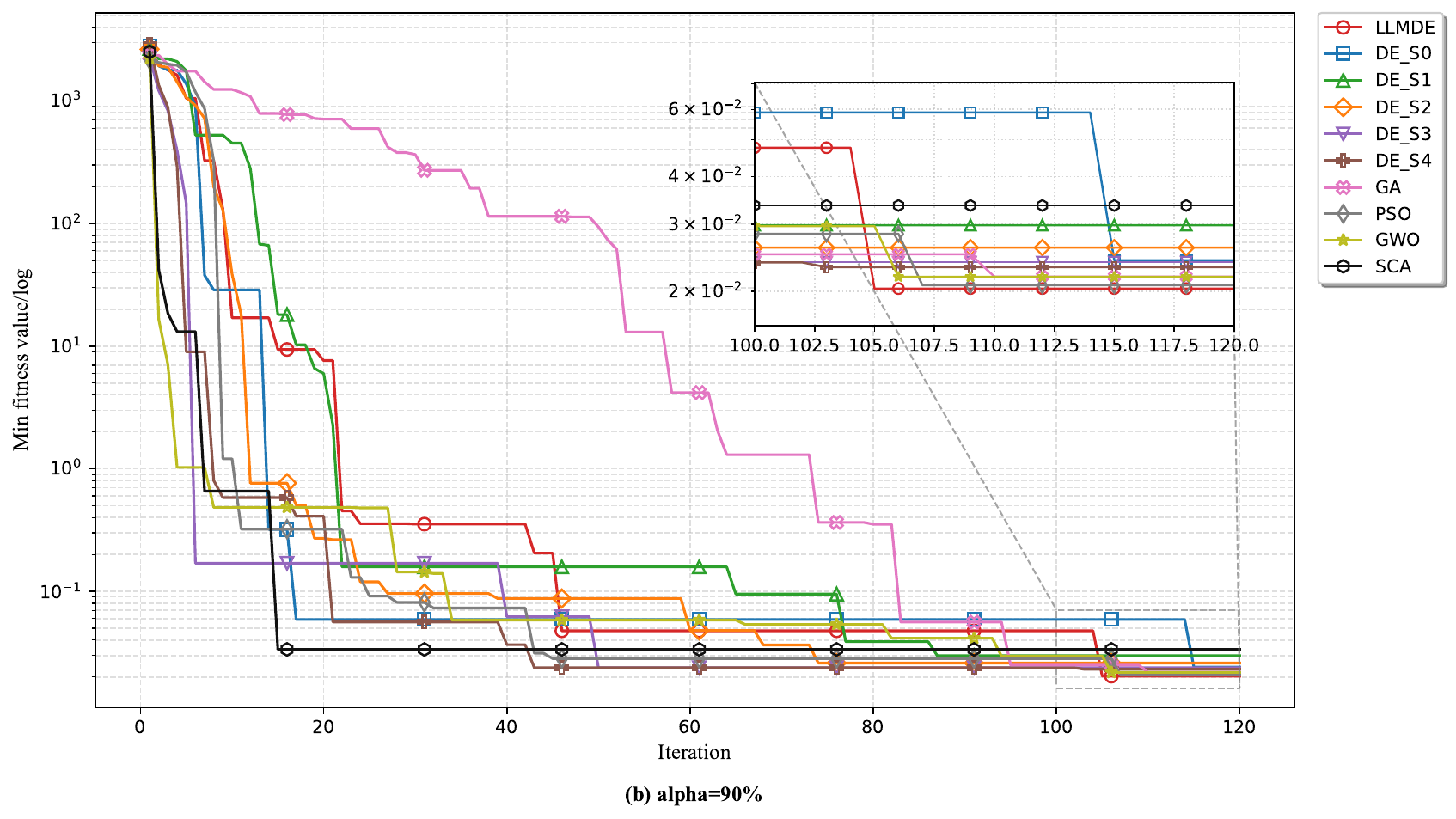}
    
    \vspace{0.5cm} 
    
    \includegraphics[width=0.48\textwidth]{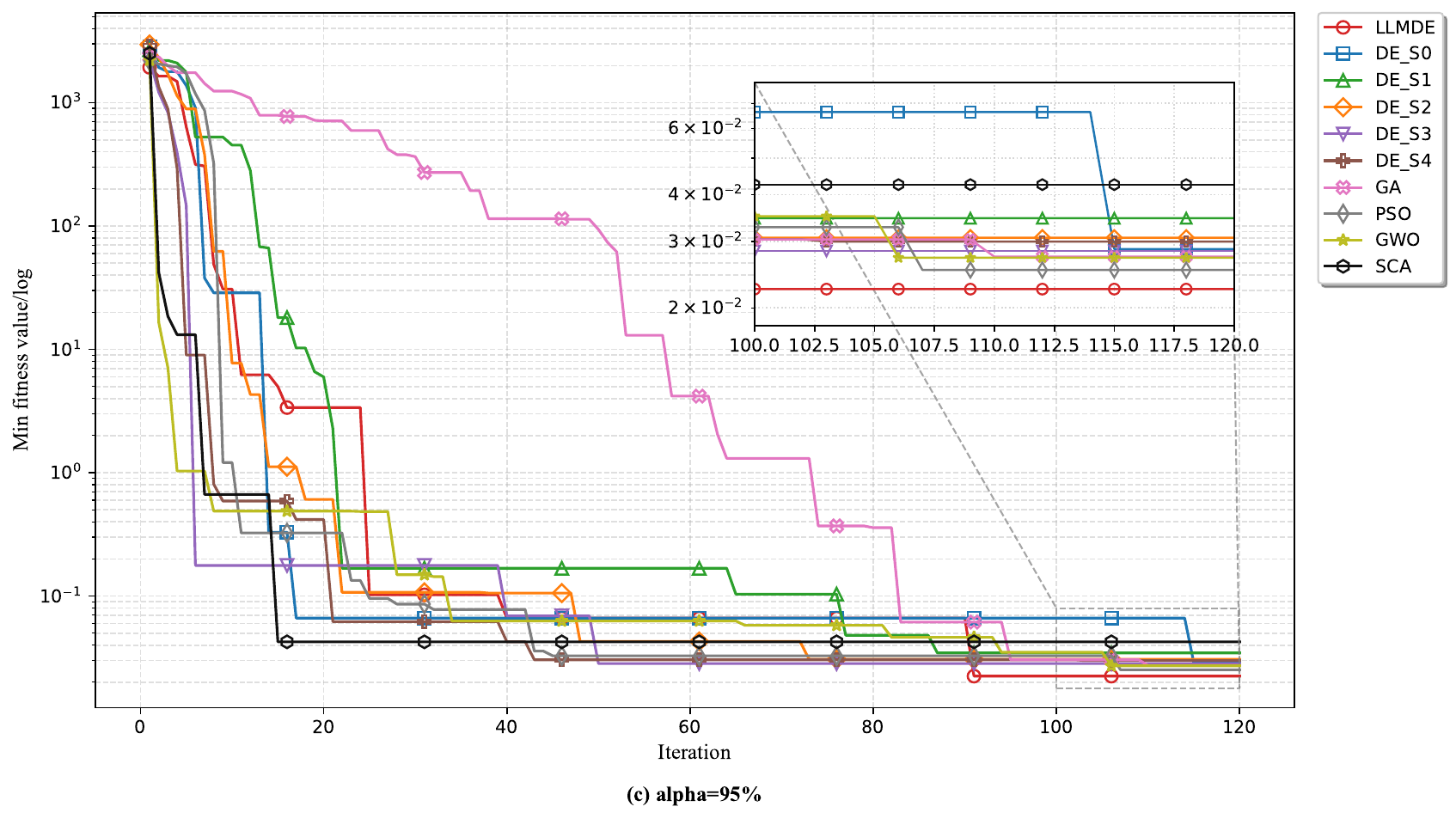}\hfill
    \includegraphics[width=0.48\textwidth]{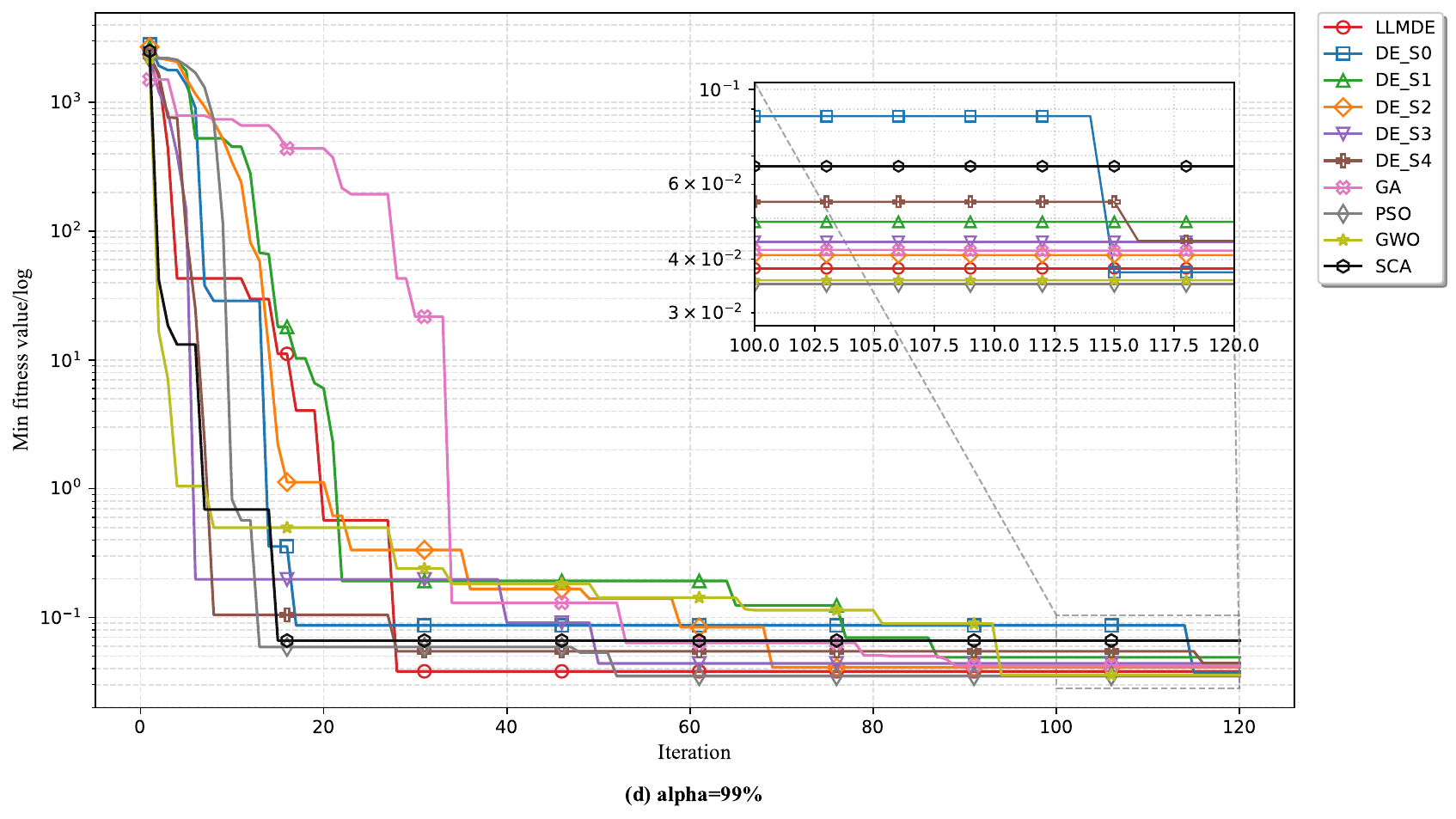} 
    
    \caption{Convergence curves under multiple confidence levels.}
    \label{Fig4}
\end{figure*}

To further inspect the search dynamics, Figure~\ref{Fig4} illustrates the convergence curves of each algorithm. As observed, LLMDE consistently improves the objective value throughout the optimization process, ultimately converging to superior solutions compared to its peers. Additionally, Table~\ref{Tab7} lists the corresponding optimal portfolio weights obtained by LLMDE. As the confidence level increases, the portfolio composition undergoes substantial reallocation, demonstrating that LLMDE can adaptively tailor asset allocation to accommodate varying risk tolerances.

\begin{table}[htbp]
    \centering
    \caption{Comparison of minimum CVaR results across confidence levels.}
    \renewcommand{\arraystretch}{1}  
    \footnotesize
    \setlength{\tabcolsep}{3pt} 
    \label{Tab9}
    \hspace*{-0.5cm}
    \begin{tabular}{cccccccccccc}
    \toprule
    \textbf{Confidence Level} & \textbf{LLMDE} & \textbf{DE\_S0} & \textbf{DE\_S1} & \textbf{DE\_S2} & \textbf{DE\_S3} & \textbf{DE\_S4} & \textbf{GA} & \textbf{PSO} & \textbf{GWO} & \textbf{SCA} \\
    \midrule
    0.75    & \cellcolor[rgb]{ .992,  .996,  .808}\textbf{0.0124 } & 0.0156  & 0.0208  & 0.0171  & 0.0167  & 0.0144  & 0.0140  & 0.0133  & 0.0144  & 0.0235  \\
    0.90    & \cellcolor[rgb]{ .992,  .996,  .808}\textbf{0.0203 } & 0.0241  & 0.0298  & 0.0260  & 0.0239  & 0.0231  & 0.0219  & 0.0207  & 0.0218  & 0.0336  \\
    0.95    & \cellcolor[rgb]{ .992,  .996,  .808}\textbf{0.0224 } & 0.0287  & 0.0346  & 0.0307  & 0.0283  & 0.0300  & 0.0273  & 0.0252  & 0.0272  & 0.0425  \\
    0.99    & 0.0381  & 0.0373  & 0.0490  & 0.0409  & 0.0439  & 0.0442  & 0.0419  & \cellcolor[rgb]{ .992,  .996,  .808}\textbf{0.0350 } & 0.0357  & 0.0661  \\
    \bottomrule
    \end{tabular}
\end{table}

\begin{table}[h] 
    \centering
    \scriptsize
    \renewcommand{\arraystretch}{0.9}
    \setlength{\tabcolsep}{4pt} 
    \caption{Optimal portfolio weights under different confidence levels.} 
    \label{Tab7}
    \begin{tabular}{lcccc} 
    \toprule
    \multirow{2}[0]{*}{\textbf{Ticker}}& \multicolumn{4}{c}{\textbf{Investment Weight}} \\
\cmidrule(lr){2-5}           & $\alpha = 75\%$ & $\alpha = 90\%$ & $\alpha = 95\%$ & $\alpha = 99\%$ \\
    \midrule
    FIX   & 0.0586  & 0.0097  & 0.0000  & 0.0251  \\
    GWW   & 0.1459  & 0.0487  & 0.0988  & 0.0658  \\
    LLY   & 0.0881  & 0.1017  & 0.1427  & 0.0231  \\
    META  & 0.0823  & 0.0264  & 0.0371  & 0.0000  \\
    MNST  & 0.1623  & 0.2154  & 0.3957  & 0.2969  \\
    MSFT  & 0.0476  & 0.2349  & 0.1580  & 0.1510  \\
    NEM   & 0.0736  & 0.0940  & 0.0000  & 0.2516  \\
    NVDA  & 0.0509  & 0.1657  & 0.0000  & 0.0062  \\
    NVR   & 0.1765  & 0.0667  & 0.1677  & 0.1802  \\
    TPL   & 0.0056  & 0.0000  & 0.0000  & 0.0000  \\
    TROW  & 0.0512  & 0.0092  & 0.0000  & 0.0000  \\
    URI   & 0.0574  & 0.0276  & 0.0000  & 0.0000  \\
    \midrule
    $CVaR_{\alpha}$  & 0.0124  & 0.0203  & 0.0224  & 0.0381  \\
    \bottomrule
    \end{tabular}
\end{table}

\section{Conclusion} 

This paper presents a novel Large Language Model-driven Differential Evolution (LLMDE) algorithm designed to address complex continuous optimization problems. By embedding an LLM into the evolutionary process as an intelligent reasoning component, the proposed framework adaptively selects mutation strategies and tunes control parameters based on real-time optimization feedback. Unlike conventional DE variants that depend on manually designed adaptation rules, LLMDE leverages the contextual understanding and reasoning capabilities of LLMs to dynamically regulate search behaviors, thereby significantly enhancing search flexibility and intelligence. Extensive evaluations on both the CEC2022 benchmark suite and the real-world CVaR portfolio optimization problem demonstrate that LLMDE consistently delivers superior optimization performance. Crucially, the proposed framework exhibits high extensibility. Since the LLM module is independent of the underlying optimizer, the framework can be naturally generalized to other metaheuristics that involve operator selection and parameter adaptation, offering a compelling paradigm for integrating LLMs with metaheuristics.

Despite its promising performance, several limitations of LLMDE warrant further investigation. First, the efficacy of LLMDE remains sensitive to prompt design and the representation of the optimization status supplied to the LLM. Constructing prompts that succinctly capture essential state information while guaranteeing consistent and robust decision-making remains a non-trivial challenge. Second, frequent LLM queries entail substantial computational overhead and API latency, which may restrict practical adoption in time-sensitive or resource-limited scenarios. Future research should prioritize reducing query frequency and computational cost, as well as evaluating alternative competitive models. Finally, the decision-making mechanism of the LLM exhibits an inherent black-box nature, obscuring the exact logic behind specific operator selections and parameter configurations across different evolutionary stages. Unraveling these underlying mechanisms will be crucial for enhancing the interpretability of LLM-assisted evolutionary algorithms.

Future work may extend LLMDE along several promising avenues. Methodologically, LLM control can be expanded to core evolutionary components such as population management, crossover design, constraint handling, and diversity preservation. To foster collaborative decision-making, employing multiple LLMs or diverse prompting strategies can boost search diversity and robustness. To enhance computational efficiency, utilizing lightweight models, Retrieval-Augmented Generation (RAG) \cite{lewis2020retrieval,zhao2026retrieval}, and historical memory mechanisms can effectively minimize latency while refining decision quality. From a practical application perspective, the framework can be adapted to multi-objective portfolio optimization, dynamic rebalancing, and scenario-based allocation under real-world trading constraints. Overall, LLM-driven evolutionary optimization represents a compelling paradigm for advancing next-generation intelligent decision-making.

\section*{CRediT authorship contribution statement}

\textbf{Rong Chai}: Methodology, Data Curation, Software, Visualization, Writing – original draft.  
\textbf{V\'{a}clav Sn\'{a}\v{s}el}: Conceptualization, Methodology, Writing-review \& editing.
\textbf{Xiaopeng Wang}: Methodology, Software, Writing-review \& editing. 
\textbf{Seyedali Mirjalili}: Methodology, Visualization, Writing-review \& editing.
\textbf{Crina Grosan}: Methodology, Writing-review \& editing. 

\section*{Declaration of competing interest}

The authors declare that they have no known competing financial interests or personal relationships that may have influenced the work reported in this paper.

\section*{Acknowledgment}

The authors gratefully acknowledge the financial support from the CLARA project funded by the European Union’s HORIZON EUROPE programme (No. 101136607).

\bibliographystyle{elsarticle-num}
\bibliography{mybibfile}

\end{document}